\documentclass[journal]{IEEEtran}
\usepackage{blindtext}
\usepackage{graphicx}

\usepackage[lofdepth,lotdepth,position=top]{subfig} % \subfloat command
\usepackage{notations}
\usepackage[normalem]{ulem}
\usepackage{float}
\usepackage{xr}
\begin{document}
%
% paper title Please 
% can use linebreaks \\ within to get better formatting as desired
%\title{Iterative Smoothing Proximal Gradient for Regression with Structured Sparsity}
%\title{High Dimensional Linear Model with Structured Sparsity Minimized with Continuation of Proximal Gradient Smoothing}
% Chen 2012 title was: SMOOTHING PROXIMAL GRADIENT METHOD FOR GENERAL STRUCTURED SPARSE REGRESSION
% Key words: 
% - Continuation of Nesterov's Smoothing 
% - High Dimensional Linear Model with Structured Sparsity
% - in Neuroimaging
%\title{Continuation of Nesterov's Smoothing to solve Linear Models with Structured Sparsity for High-Dimensional Neuroimaging}
% or
%\title{Linear Models with Structured Sparsity for High Dimensional Neuroimaging Minimized using Continuation of Nesterov's Smoothing}
%\title{Continuation of Nesterov's Smoothing when solving Linear Regression Problems with Structured Sparsity in High-Dimensional Neuroimaging}
\title{Continuation of Nesterov's Smoothing for Regression with Structured Sparsity in High-Dimensional Neuroimaging}

%
%
% author names and IEEE memberships
% note positions of commas and nonbreaking spaces ( ~ ) LaTeX will not break
% a structure at a ~ so this keeps an author's name from being broken across
% two lines.
% use \thanks{} to gain access to the first footnote area
% a separate \thanks must be used for each paragraph as LaTeX2e's \thanks
% was not built to handle multiple paragraphs
%

\author{
Fouad Hadj-Selem\textsuperscript{*},
Tommy L\"ofstedt\textsuperscript{*},
Elvis Dohmatob,
Vincent Frouin,
Mathieu Dubois,
Vincent Guillemot,
and Edouard Duchesnay\textsuperscript{*}
%and Alzheimer's Disease Neuroimaging Initiative
\thanks{E. Duchesnay, V. Frouin, V. Guillemot and M. Dubois are with NeuroSpin, CEA, Universit\'e Paris-Saclay - France.}% <-this % stops a space
\thanks{F. Hadj-Selem is with the Energy Transition Institute: VeDeCoM - France.}% <-this % stops a space
\thanks{T. L\"ofstedt is with the Department of Radiation Sciences, Ume\aa{} University, Ume\aa{} - Sweden.}
\thanks{E. Dohmatob is with PARIETAL Team, INRIA / CEA, Universit\'e Paris-Saclay - France.}% <-this % stops a space
%\thanks{Alzheimer's Disease Neuroimaging Initiative: Data used in the preparation of this article were obtained from the Alzheimer's Disease Neuroimaging Initiative (ADNI) database (http://www.loni.ucla.edu/ADNI). As such, the investigators within the ADNI contributed to the design and implementation of ADNI and/or provided data but did not participate in analysis or writing of this report. ADNI investigators include (complete listing available at http://www.loni.ucla.edu/ADNI/Collaboration/ADNI Author ship list.pdf)}
\thanks{\textsuperscript{*}Contributed equally.}
%\thanks{Supported by grants from the French National Research Agency: ANR IA BRAINOMICS (ANR-10-BINF-04) and a European Commission grant: MESCOG (FP6 ERA-NET NEURON 01 EW1207).}
}

% note the % following the last \IEEEmembership and also \thanks - 
% these prevent an unwanted space from occurring between the last author name
% and the end of the author line. i.e., if you had this:
% 
% \author{....lastname \thanks{...} \thanks{...} }
%                     ^------------^------------^----Do not want these spaces!
%
% a space would be appended to the last name and could cause every name on that
% line to be shifted left slightly. This is one of those "LaTeX things". For
% instance, "\textbf{A} \textbf{B}" will typeset as "A B" not "AB". To get
% "AB" then you have to do: "\textbf{A}\textbf{B}"
% \thanks is no different in this regard, so shield the last } of each \thanks
% that ends a line with a % and do not let a space in before the next \thanks.
% Spaces after \IEEEmembership other than the last one are OK (and needed) as
% you are supposed to have spaces between the names. For what it is worth,
% this is a minor point as most people would not even notice if the said evil
% space somehow managed to creep in.

% The paper headers
\markboth{IEEE Transactions on Medical Imaging}%,~Vol.~6, No.~1, January~2007}%
{Hadj-Selem \MakeLowercase{\textit{et al.}}: Bare Demo of IEEEtran.cls for Journals}

% The only time the second header will appear is for the odd numbered pages
% after the title page when using the twoside option.
% 
% *** Note that you probably will NOT want to include the author's ***
% *** name in the headers of peer review papers.                   ***
% You can use \ifCLASSOPTIONpeerreview for conditional compilation here if
% you desire.

% If you want to put a publisher's ID mark on the page you can do it like
% this:
%\IEEEpubid{0000--0000/00\$00.00~\copyright~2007 IEEE}
% Remember, if you use this you must call \IEEEpubidadjcol in the second
% column for its text to clear the IEEEpubid mark.

% make the title area
\maketitle

\begin{abstract}
Predictive models can be used on high-dimensional brain images to decode cognitive states or diagnosis/prognosis of a clinical condition/evolution.
Spatial regularization through structured sparsity offers new perspectives in this context and reduces the risk of overfitting the model while providing interpretable neuroimaging signatures by forcing the solution to adhere to domain-specific constraints.
Total Variation (TV) is a promising candidate for structured penalization: it enforces spatial smoothness of the solution while segmenting predictive regions from the background.
We consider the problem of minimizing the sum of a smooth convex loss, a non-smooth convex penalty (whose proximal operator is known) and a wide range of possible complex, non-smooth convex structured penalties such as TV or overlapping group Lasso.
Existing solvers are either limited in the functions they can minimize or in their practical capacity to scale to high-dimensional imaging data.
Nesterov's smoothing technique can be used to minimize a large number of non-smooth convex structured penalties.
However, reasonable precision requires a small smoothing parameter, which slows down the convergence speed to unacceptable levels.
To benefit from the versatility of Nesterov's smoothing technique, we propose a first order continuation algorithm, CONESTA, which automatically generates a sequence of decreasing smoothing parameters.
The generated sequence maintains the optimal convergence speed towards any globally desired precision.
Our main contributions are:
To propose an expression of the duality gap to probe the current distance to the global optimum in order to adapt the smoothing parameter and the convergence speed.
This expression is applicable to many penalties and can be used with other solvers than CONESTA.
We also propose an expression for the particular smoothing parameter that minimizes the number of iterations required to reach a given precision.
Further, we provide a convergence proof and its rate, which is an improvement over classical proximal gradient smoothing methods.
We demonstrate on both simulated and high-dimensional structural neuroimaging data that CONESTA significantly outperforms many state-of-the-art solvers in regard to convergence speed and precision.
\end{abstract}
% IEEEtran.cls defaults to using nonbold math in the Abstract.
% This preserves the distinction between vectors and scalars. However,
% if the journal you are submitting to favors bold math in the abstract,
% then you can use LaTeX's standard command \boldmath at the very start
% of the abstract to achieve this. Many IEEE journals frown on math
% in the abstract anyway.

% Note that keywords are not normally used for peerreview papers.

% For peer review papers, you can put extra information on the cover
% page as needed:
% \ifCLASSOPTIONpeerreview
% \begin{center} \bfseries EDICS Category: 3-BBND \end{center}
% \fi
%
% For peerreview papers, this IEEEtran command inserts a page break and
% creates the second title. It will be ignored for other modes.
\IEEEpeerreviewmaketitle

%%%%%%%%%%%%%%%%%%%%%%%%%%%%%%%%%%%%%%%%%%%%%%%%%%%%%%%%%%%%%%%%%%%%%%%%%%%%%%%%
\section{Introduction}\label{sec:introducing}
%%%%%%%%%%%%%%%%%%%%%%%%%%%%%%%%%%%%%%%%%%%%%%%%%%%%%%%%%%%%%%%%%%%%%%%%%%%%%%%%

\FIRST{
Multivariate machine learning (ML) applied in neuroimaging offers new perspectives in early diagnosis and prognosis of brain diseases~\cite{arbabshirani_single_2016} using different MRI modalities, as well as brain decoding in cognitive neurosciences~\cite{hanes_ed._john-dylan_multivariate_2011} using functional MRI.
Indeed, ML algorithms can evaluate a vast number of brain-related features and capture complex relationships in the data with the purpose of making inferences at a single-subject level.
The learning and application of ML algorithms should address the risk of overfitting when applied to high-dimensional training data; while simultaneously achieving their goal of providing predictive and interpretable brain patterns.
Additionally, these predictive brain patterns should be stable with regard to some variability of the training subjects with similar clinical and demographical characteristics.

Issues regarding increased interpretability and high dimensionality can be addressed simultaneously using} \SCND{\textit{e.g.}} \FIRST{a limited ($\leq 100$) number of atlas-based regions of interest (ROI) where the signal within each ROI is averaged.
In a comparison study~\cite{cuingnet_classif_adni_2011}, voxel-based approaches outperformed region-based approaches, suggesting that the benefits obtained by dimensionality reduction do not compensate for the loss of informative signal caused by averaging within pre-defined regions.
Indeed, regional approaches rely on the assumption that the spatial extent of the unknown predictive brain region matches with one of the atlas-based ROIs.
Hence, this implies that ML algorithms should perform a data-driven selection of features that are organized in a few regions while still working at the voxel level.

% Spatial regularization
This objective can be addressed with spatial regularization that simultaneously reduces the risk of overfitting while enforcing the solution to be organized in interpretable regions of connected voxels.}
Such regularizations extend classical predictors by introducing penalties that enforce local smoothness on the model parameters.
%Models with spatial regularization were recently proposed as means to reduce the risk of overfitting, while simultaneously improving the interpretability of the solution. Such regularization extends classical predictors by introducing penalties that enforce local smoothness on the model parameters.
Graph-Net~\cite{grosenick_interpretable_2013, dohmatob_speeding-up_2015} forces adjacent voxels to have similar weights using a squared $\ell_2$ penalty on the spatial gradients of the weight map.
This penalty is differentiable, which considerably simplifies the optimization problem. However, Graph-Net results in a smooth map without clearly identified regions of piece-wise constant clusters of related voxels.
% TV
The \FIRST{Total Variation (TV)} penalty forces sparsity on the spatial derivatives of the weight map (using an $\ell_{12}$-norm), segmenting the weight map into spatially-contiguous parcels with almost constant values~\cite{michel2011}.
TV can be combined with sparsity-inducing penalties (such as \FIRST{$\ell_1$ (Lasso)}~\cite{tibshirani96}) to obtain segmenting properties that extracts predictive regions from a noisy background with zeros~\cite{gramfort_identifying_2013, dubois_predictive_2014, dohmatob_benchmarking_l1tv_2014}.
TV, together with Lasso, produces the desired foreground-background segmentation by imposing constant-valued parcels.

We propose a solver that addresses a very general class of optimization problems including many group-wise penalties (allowing overlapping groups) such as Group Lasso and TV. The function that we wish to minimize has the form
\FIRST{
\begin{align}\label{eq:optimization_problem}
    % \min_{\bbeta} f
    f(\bbeta)
    %\equiv
    =
    &~
    \overbrace{g(\bbeta)}^{\text{smooth}}
    +
    \overbrace{\loneCoeff h(\bbeta) + \structCoeff s(\bbeta)}^{\text{non-smooth}},
\end{align}
where $\bbeta$ is the vector of parameters to be estimated. $g$ is the sum of a differentiable loss $\Loss$ with any differentiable penalty, \eg, the least-squares loss with the $\ell_2$ (ridge) penalty: $g(\bbeta) = \frac{1}{2}\|\X \bbeta - \y\|_2^2 + \frac{\ltwoCoeff}{2} \|\bbeta\|_2^2$.
The $h$ is a penalty whose proximal operator is known, \eg, the $\ell_1$ penalty: $h(\bbeta) = \norm{1}{\bbeta}$. This regularization will induce sparsity on the parameter map.
The $s$ is a complex penalty on the structure of the input variables, for which the proximal operator is either not known, or for which the proximal operator is too expensive to compute.
The only assumption is that $s$ is an $\ell_{12}$ group norm of the following form:
% TV: the $\ell_1$ norm of the image gradient
\begin{equation}\label{eq:s_groupnorm}
s(\bbeta) = \sum_{\grp \in \setGrp} \| \A_{\grp} \bbeta_\grp\|_2,
\end{equation}
% \label{eq:optimization_problem}
where $\setGrp = \{\grp_1, \cdots, \grp_{|\setGrp|}\}$ denotes the set of, possibly overlapping groups of features and $\A_{\grp}$ any linear operator on each group.
Overlapping group Lasso and TV belong to this class of functions.
The contribution of the penalties is controlled by the scalar weights: $\loneCoeff, \ltwoCoeff$ and $\structCoeff $.
}

Although many solvers have already been proposed that can minimize this function, their practical use in the context of high-dimensional neuroimaging data ($\geq 10^5$ features) remains an open issue (see \suppRef{supp:sec:state-of-the-art-solvers:background} and \cite{faasta, dohmatob_benchmarking_l1tv_2014}). This paper compares the convergence speed of the proposed solver to that of four state-of-the-art solvers: 
(i) An \textit{Inexact} proximal gradient algorithm (Inexact FISTA)~\cite{Schmidt2011}, where the proximal operator of $s$ is approximated numerically. While the main algorithm (FISTA) enjoys a convergence rate of $\bigO{1/k^2}$ (with $k$ the number of iterations), the precision of the approximation of the proximal operator is required to decrease as $\bigO{1 / k^{4+\delta}}$ for any $\delta>0$~\cite[Proposition~2]{Schmidt2011}.
This results in prohibitive computations in order to reach a reasonable precision, especially with high-dimensional $\bbeta$ vectors as found with brain images that generally involve $\geq 10^4$ (functional MRI) to $\geq 10^5$ (structural MRI) features.
(ii) The Excessive Gap Method (EGM)~\cite{Nesterov2005}, which does not allow true sparsity nor general loss functions.
(iii) The alternating direction method of multipliers (ADMM)~\cite{Boyd_etal_2011}, which \FIRST{is computationally expensive and/or difficult to compute for general structured penalties and which} suffers from a difficulty of setting the regularization parameter, as mentioned in~\cite{dohmatob_benchmarking_l1tv_2014, Schmidt2011}.
(iv) Nesterov's smoothing technique~\cite{Nesterov2005}, which provides an appealing and generic framework in which a large range of non-smooth convex structured penalties can be minimized without computing their proximal operators. However, reasonable precision ($\approx 10^{-3}$ or higher) requires a very small smoothing parameter, which brings down the convergence rate to unacceptable levels.

We propose a continuation solver, called \textit{CONESTA} (short for \textbf{CO}ntinuation with \textbf{NE}sterov smoothing in a \textbf{S}hrinkage-\textbf{T}hresholding \textbf{A}lgorithm), based on Nesterov's smoothing technique, that automatically generates a decreasing sequence of smoothing parameters in order to maintain the optimal convergence speed towards any globally desired precision.
Nesterov's smoothing technique makes the solver highly versatile: it can address a large class of complex penalties (the function $s$ in \eqref{eq:optimization_problem}) where the proximal operators are either not known or expensive to compute.
The problem can be minimized using an accelerated proximal gradient method, possibly also utilizing (non-smoothed, \eg \FIRST{$\ell_1$}) sparsity-inducing penalties.
CONESTA can be understood as a smooth touchdown procedure that uses the duality gap to probe the distance to the ground (global optimum) and dynamically adapts its speed (the smoothing parameter) according to this distance.

As a first contribution, we propose an expression of the duality gap to control the convergence of the global non-smooth problem.
This expression is applicable to a large range of structured penalties, which makes it useful beyond the scope of the present paper.
%proposed CONESTA algorithm.
The duality gap provides an upper bound on the current error, which can be used to derive the next target precision to be reached by the smoothed problem.

As a second contribution, we propose an expression that provides the particular smoothing parameter that minimizes the number of iterations required to reach the above-mentioned prescribed precision.

As a third contribution, we provide a convergence proof and rate, which is an improvement over classical (without continuation) proximal gradient smoothing.

The fourth contribution is the experimental demonstration, on both simulated and high-dimensional structural neuroimaging data, that CONESTA significantly outperforms the state-of-the-art solvers, \ie, EGM, ADMM, classical FISTA with fixed smoothing and Inexact FISTA, in regards to convergence speed and/or precision of the solution.
We further demonstrate the relevance of TV for extracting stable predictive signatures from structural MRI data.

%% -------------------------------------------------------------------------- %%
\subsection{Reformulating TV as a linear operator} \label{sec:tv}
%% -------------------------------------------------------------------------- %%

Before discussing the minimization with respect to $\bbeta$, we provide details on the
encoding of the spatial structure within the penalty $s$.
It is essential to note that the algorithm is independent of the spatial structure of the data.
All the structural information is encoded in the linear operator, $\A$, and it is computed outside of the algorithm.
Thus, the algorithm has the ability to address various structured data \FIRST{(arrays, meshes)} and, most importantly, also other penalties aside from the TV penalty.
Using our generic CONESTA solver for any penalty, which can be written in the form of \eqref{eq:optimization_problem}, requires only building the linear operator $\A$; and a projection function, detailed in \eqref{eq:projection_l2}.
This section presents the formulation and the design of $\A$ in the specific case of the TV penalty applied to the parameter vector $\bbeta$ measured on a 3-dimensional~(3D) image.
%or a mesh of the cortical surface\comment{We do not present it for the mesh. Remove or add to supplementary?}.

%3d->1d mapping
A brain mask is used to establish a mapping, $\phi(i, j, k)$, from integer coordinates $(i, j, k)$ in the 3D grid of the brain image, and an index $\phi \in \{1,\ldots,\nDim\}$ in the collapsed (vectorized) image. We extract the spatial (forward) neighborhood at $(i, j, k)$, of size $\leq 4$, corresponding to a voxel and its three neighboring voxels, within the mask, in the positive $i$, $j$ and $k$ directions. The TV penalty is defined as
\begin{equation} \label{eq:tv_def}
    \TV(\bbeta) \equiv \sum_{i,j,k}
        \Big\|\nabla\big(\bbeta_{\phi(i,j,k)}\big)\Big\|_2,
\end{equation}
where $\nabla ( \bbeta_{\phi(i, j, k)} )$ denotes the spatial gradient of the parameter map, $\bbeta\in\mathbb{R}^\nDim$, at the 3D position $(i, j, k)$ mapped to element $\phi(i, j, k)$ in $\bbeta$. A first order approximation of the spatial gradient, $\nabla(\bbeta_{\phi(i, j, k)})$, can be computed by applying the linear operator $\A_\phi' \in \mathbb{R}^{3 \times 4}$ to the parameter vector $\bbeta'_{\phi(i,j,k)} \in \mathbb{R}^{4}$ as
\begin{equation} \label{eq:grad_Ag}
\nabla\left( \bbeta_{\phi(i, j, k)} \right) \FIRST{\equiv}
    \underbrace{
        \left[
            \begin{array}{cccc}
                -1 & 1 & 0 & 0 \\ 
                -1 & 0 & 1 & 0 \\
                -1 & 0 & 0 & 1
            \end{array}
        \right]%
        \vphantom{\left[\begin{array}{l}\bbeta_{\phi(i,j,k)}\\\bbeta_{\phi(i+1,j,k)}\\\bbeta_{\phi(i,j+1,k)}\\\bbeta_{\phi(i,j,k+1)}\end{array}\right]}%
    }_{\A'_\phi}
    \underbrace{
        \left[
            \begin{array}{l}
                \bbeta_{\phi(i,j,k)}\\
                \bbeta_{\phi(i+1,j,k)}\\
                \bbeta_{\phi(i,j+1,k)}\\
                \bbeta_{\phi(i,j,k+1)}
            \end{array}
        \right]
    }_{\bbeta'_{\phi(i,j,k)}},
\end{equation}
where thus $\bbeta'_{\phi(i,j,k)}$ contains the elements at linear indices in the collapsed parameter map, $\bbeta$, corresponding to the spatial neighborhood in the 3D image at $\bbeta_{\phi(i,j,k)}$.
Then, $\A'_\phi$ is extended, by zeros, to a large but very sparse matrix $\A_{\phi(i,j,k)} \in \mathbb{R}^{3 \times \nDim}$ such that $\A'_\phi\bbeta'_{\phi(i,j,k)} = \A_{\phi(i,j,k)}\bbeta$. If some neighbors lie outside of the mask, the corresponding rows in $\A_{\phi(i,j,k)}$ are removed \FIRST{(or set to zero)}. Approximating TV by
\begin{equation}\label{eq:s_asl12}
    \TV(\bbeta) \FIRST{=} \sum_{i,j,k} \| \A_{\phi(i,j,k)} \bbeta \|_2
\end{equation}
allows us to use the TV, \FIRST{as the structured penalty $s$}, in \eqref{eq:optimization_problem}.
Finally, with a vertical concatenation of all the $\A_{\phi(i,j,k)}$ matrices, we obtain the full linear operator $\A \in \mathbb{R}^{3\nDim \times \nDim}$ that will be used in \FIRST{the following sections}.

%%%%%%%%%%%%%%%%%%%%%%%%%%%%%%%%%%%%%%%%%%%%%%%%%%%%%%%%%%%%%%%%%%%%%%%%%%%%%%%%
\section{Nesterov's smoothing of the structured penalty} \label{sec:Nesterov}
%%%%%%%%%%%%%%%%%%%%%%%%%%%%%%%%%%%%%%%%%%%%%%%%%%%%%%%%%%%%%%%%%%%%%%%%%%%%%%%%

We consider the convex non-smooth minimization of \eqref{eq:optimization_problem}
with respect to $\bbeta$. This problem includes a general
structured penalty, $s$, that (for the purpose of this paper) covers the specific case of TV.
The accelerated proximal gradient algorithm (FISTA)~\cite{Beck2009} can be used to solve the problem when applying only \eg the $\ell_1$ penalty.
%A widely used approach when dealing with non-smooth problems is to use methods based on the proximal operator of the penalties. For the $\ell_1$ penalty alone, the proximal operator is analytically known and being solved with ISTA~\cite{Daubechies2004} or FISTA~\cite{Beck2009}.
However, since the proximal operator of TV, together with the $\ell_1$ penalty, has no closed-form expression, standard implementations of those algorithms are not suitable. In order to overcome this barrier we used Nesterov's smoothing technique~\cite{Nesterov2005a}, which consists of approximating the non-smooth penalty for which the proximal operator is unknown (\eg, TV) with a smooth function (for which the gradient is known). Non-smooth penalties with known proximal operators (\eg, $\ell_1$) are not affected by this smoothing. Hence, as described in~\cite{Chen2012}, this allowed us to use an exact accelerated proximal gradient algorithm.

Using the dual norm of the $\ell_2$-norm (\ie the $\ell_2$-norm), \eqref{eq:s_asl12} can be reformulated as
\begin{align} \label{eq:s_asl1_dual}
    \TV(\bbeta) &\FIRST{=} %s(\bbeta) \nonumber\\
                %&=
                \sum_{i,j,k} \| \A_{\phi(i,j,k)} \bbeta \|_2 \nonumber\\
                &\FIRST{=} \sum_{i,j,k} \max_{\|\balpha_{\phi(i,j,k)}\|_2\leq 1} \tran{\balpha}_{\phi(i,j,k)} \A_{\phi(i,j,k)} \bbeta,
\end{align}
where $\balpha_{\phi(i,j,k)} \in \setK_{\phi(i,j,k)} = \{\balpha_{\phi(i,j,k)} \in \mathbb{R}^{3}: \|\balpha_{\phi(i,j,k)}\|_2\leq 1\}$ is a vector of auxiliary variables in the $\ell_2$ unit ball, associated with $\A_{\phi(i,j,k)} \bbeta$.
As with $\A \in \mathbb{R}^{3\nDim \times \nDim}$, which is the vertical concatenation of all the $\A_{\phi(i,j,k)}$, we concatenate all the $\balpha_{\phi(i,j,k)}$ to form $\balpha \in \setK=\{[\balpha_1^{\T}, \ldots, \balpha_{\nDim}^{\T}]^{\T}: \balpha_l \in \setK_l,\, \forall\, l={\phi(i,j,k)}\in\{1,\ldots,\nDim\}\} \in \mathbb{R}^{3\nDim}$.
The set $\setK$ is the Cartesian product of closed 3D unit balls in Euclidean space and, therefore, a compact convex set. \eqref{eq:s_asl1_dual} can now further be written as
\begin{equation}\label{eq:s_dual}
   \TV(\bbeta) \FIRST{=} \max_{\balpha \in \setK} \balpha^{\T} \A \bbeta
               = s(\bbeta),
\end{equation}
and with this formulation of $s$, we can apply Nesterov's smoothing technique. For a given smoothing parameter, $\mu > 0$, the function $s$ is approximated by the smooth function
\begin{equation}\label{eq:s_dual_smoothed}
   s_{\mu}(\bbeta) = \max_{\balpha \in \setK} \left\lbrace \balpha^{\T} \A \bbeta - \frac{\mu}{2}\|\balpha\|_2^2 \right\rbrace,
\end{equation}
for which $\lim_{\mu \rightarrow 0} s_{\mu}(\bbeta) = s(\bbeta)$. Nesterov~\cite{Nesterov2005a} demonstrates this convergence using the inequality in \eqref{eq:nesterov_inequality}.
The value of $\balpha_\mu^{*}(\bbeta) = [\balpha_{\mu, 1}^{*\T}, \ldots, \balpha_{\mu, \phi(i,j,k)}^{*\T}, \ldots ,\balpha_{\mu, \nDim}^{*\T}]^{\T}$ that maximizes \eqref{eq:s_dual_smoothed} is the concatenation of projections of the vectors $\A_{\phi(i,j,k)}\bbeta \in \mathbb{R}^{3}$ onto the $\ell_2$ ball $\setK_{\phi(i,j,k)}$, \ie $\balpha^*_{\mu, {\phi(i,j,k)}}(\bbeta) = \mathrm{proj}_{\setK_{\phi(i,j,k)}}\left( \frac{\A_{\phi(i,j,k)}\bbeta}{\mu} \right)$, where
\begin{equation}\label{eq:projection_l2}
 \mathrm{proj}_{\setK_{\phi(i,j,k)}}(\vector{x}) = \begin{cases}
                            \vector{x} & \text{~if~} \|\vector{x}\|_2 \leq 1\\ 
                            \frac{\vector{x}}{\|\vector{x}\|_2} & \text{~otherwise.} \\
                        \end{cases}
\end{equation}

%% Gradient
The function $s_\mu$, \ie Nesterov's smooth transform of $s$, is convex and differentiable. Its gradient is given by Nesterov~\cite{Nesterov2005a} as
\begin{equation} \label{eq:gradient_nesterov}
    \nabla s_{\mu}(\bbeta) = \A^{\T}\balpha^*_{\mu}(\bbeta).
\end{equation}
The gradient is Lipschitz-continuous, with constant
\begin{equation} \label{eq:nesterov_lipschitz}
    L\big(\nabla(s_{\mu})\big) = \frac{\|\A\|_2^2}{\mu},
\end{equation}
in which \(\|\A\|_2\) is the matrix spectral norm of $\A$.
Moreover, Nesterov~\cite{Nesterov2005a} provides the following inequality, relating $s_\mu$ and $s$
\begin{equation} \label{eq:nesterov_inequality}
    s_\mu(\bbeta)\leq s(\bbeta)\leq s_\mu(\bbeta) + \mu M,\quad \forall \bbeta \in \mathbb{R}^\nDim,
\end{equation}
where $M = \max_{\balpha \in \setK} \frac{1}{2}\|\balpha\|_2^2=\frac{\nDim}{2}$.

Thus, a new (smoothed) function, closely related to \eqref{eq:optimization_problem}, arises as
\begin{gather}\label{eq:optimization_problem_smoothed}
    %\min_{\bbeta} f_{\mu}  \equiv
    f_\mu(\bbeta) =
    \overbrace{
        \underbrace{
            \Loss(\bbeta)\!
            +
            \ltwoCoeff\norm{2}{\bbeta}^2
            \vphantom{\Big(}
        }_{g(\bbeta)}
      %+ \structCoeff s_{\mu}(\bbeta)}_{\text{smooth}}
            + \structCoeff
            \underbrace{
                \left\lbrace
                    {\balpha_\mu^{*}(\bbeta)}^{\T} \A \bbeta
                    -
                    \frac{\mu}{2}\|\balpha^*\|_2^2
                \right\rbrace
            }_{s_\mu(\bbeta)}
    }^{\text{smooth}}
    +
    \loneCoeff%
    \hspace{-0.55em}
    \overbrace{%
        \underbrace{%
            \norm{1}{\bbeta}
            \vphantom{\Big(}
        }_{h(\bbeta)}
    }^{\text{non-smooth}}%
    \hspace{-0.7em}.
\end{gather}

Hence, we can explicitly compute the gradient of the smooth part, $\nabla(g+\structCoeff s_\mu)$, \eqref{eq:gradient_nesterov}, its Lipschitz constant,  \eqref{eq:nesterov_lipschitz}, and also the proximal operator of the non-smooth part. We thus have all the necessary ingredients to minimize the function using \eg an accelerated proximal gradient method~\cite{Beck2009}.
Given a starting point, $\bbeta^0$, and a smoothing parameter, $\mu$, FISTA (Algorithm~\ref{algo:fista}) minimizes the smoothed function and reaches a given precision, $\varepsilon_\mu$.

\begin{algorithm} 
  \caption{FISTA\big($\bbeta^0$, $\varepsilon_\mu$, $\mu$, $\A$, $g$, $s_\mu$, $h$, $\structCoeff$, $\loneCoeff$\big)}\label{algo:fista}
  \begin{algorithmic}[1]
  %\Require $\bbeta^0$, $\mu$, $\varepsilon_\mu$, $\A$, $\ltwoCoeff$, $\nabla(g+\structCoeff s_\mu)(.)$, $L(\nabla(g))$
      \State $\bbeta^1 = \bbeta^0$; $k=2$ 
      \State Step size $t_\mu = \left(L(\nabla(g)) + \structCoeff \frac{\|\A\|_2^2}{\mu}\right)^{-1}$ \label{algo:fista:step}%, fix the step size as in \eqref{eq:t_mu}
      \Repeat
	\State $\mathbf{z} = \bbeta^{k-1} + \frac{k-2}{k+1}\left(\bbeta^{k-1}-\bbeta^{k-2}\right)$
	\State $\bbeta^k = \prox_{\loneCoeff h}\big(\mathbf{z} - t_\mu\nabla (g+\structCoeff s_\mu) (\mathbf{z})\big)$ %TO BE CHECKED
      %\State \textbf{if}  \textbf{then break end if}
      %\State $k = k+1$      
      \Until{$\textsc{Gap}_{\mu}(\bbeta^{k}) \leq \varepsilon_\mu$} (see \secref{sec:duality_gap})\label{algo:fista:stop}
      \State \textbf{return} $\bbeta^{k}$
  \end{algorithmic}
\end{algorithm}

%%%%%%%%%%%%%%%%%%%%%%%%%%%%%%%%%%%%%%%%%%%%%%%%%%%%%%%%%%%%%%%%%%%%%%%%%%%%%%%%
\section{The CONESTA algorithm} \label{sec:conesta}
%%%%%%%%%%%%%%%%%%%%%%%%%%%%%%%%%%%%%%%%%%%%%%%%%%%%%%%%%%%%%%%%%%%%%%%%%%%%%%%%

The step size, $t_\mu$, computed in Line~\ref{algo:fista:step} of Algorithm~\ref{algo:fista}, must be smaller than or
equal to the reciprocal of the Lipschitz constant of the gradient of the smooth part, \ie of $g+\structCoeff s_\mu$~\cite{Beck2009}.
This relationship between $t_\mu$ and $\mu$ implies a trade-off between speed and precision: Indeed, the FISTA convergence rate, given in the \suppEqref{supp:eq:FISTA_convergence_upper_bound}, shows that a high precision (small $\mu$ and $t_\mu$) will lead to a slow convergence. Conversely, poor precision (large $\mu$ and $t_\mu$) will lead to rapid convergence.

To optimize this trade-off, we propose a continuation approach (Algorithm \ref{algo:conesta}) that decreases the smoothing parameter with respect to the distance to the minimum. On the one hand, when we are far from $\bbeta^*$ (the minimum of \eqref{eq:optimization_problem}), we can use a large $\mu$ to rapidly decrease the objective function. On the other hand,  when we are close to $\bbeta^*$, we need a small $\mu$ in order to obtain an accurate approximation of the original objective function.

%%----------------------------------------------------------------------------%%
\subsection{Duality gap} \label{sec:duality_gap}
%%----------------------------------------------------------------------------%%

The distance to the unknown $f(\bbeta^*)$ is estimated using a duality gap.
Duality formulations are often used to control the achieved precision level when minimizing convex functions.
The duality gap provides an upper bound of the error, $f(\bbeta^k) - f(\bbeta^*)$, for any $\bbeta^k$, when the minimum is unknown.
Moreover, it vanishes at the minimum:
\begin{equation} \label{eq:gap_prop}
    \begin{array}{rl}
        \textsc{Gap}(\bbeta^k) \geq f(\bbeta^k) - f(\bbeta^*) &\geq\;\, 0, \\
        \textsc{Gap}(\bbeta^*)                                &=\;\,    0.
    \end{array}
\end{equation}

The duality gap is the cornerstone of the CONESTA algorithm. Indeed, it is used three times:
\begin{enumerate}\renewcommand{\labelenumi}{\roman{enumi}}
    \item[(i)] As the stopping criterion in the inner FISTA loop (Line~\ref{algo:fista:stop} in Algorithm~\ref{algo:fista}). FISTA will stop as soon as the current precision is achieved using the current smoothing parameter, $\mu$. This prevents unnecessary iterations toward the approximated (smoothed) objective function.
    \item[(ii)] In the $i$th CONESTA iteration, as a way to estimate the current error $f(\bbeta^i) - f(\bbeta^*)$ (Line~\ref{algo:conesta:eps} in Algorithm~\ref{algo:conesta}).
    %The error is estimated using the gap of the smoothed problem, $\textsc{Gap}_{\mu=\mu^i}(\bbeta^{i+1})$, which avoids unnecessary computation since it has already been computed during the last iteration of FISTA.
    %The inequality in \eqref{eq:nesterov_inequality} is used to obtain the distance, $\varepsilon^i$, to the original non-smoothed problem.
    % This value is then used to deduce all the other parameters for the next application of FISTA.
    The next desired precision, $\varepsilon^{i+1}$, and the smoothing parameter, $\mu^{i+1}$ (using \theoremref{thm:optimal_mu}), are derived from this value.
    \item[(iii)] Finally, as the global stopping criterion in CONESTA (Line~\ref{algo:conesta:stop} in Algorithm~\ref{algo:conesta}).
    This guarantees that the obtained approximation of the minimum, $\bbeta^{i}$, at convergence, satisfies $f(\bbeta^{i}) - f(\bbeta^*) < \varepsilon$.
\end{enumerate}

\eqref{eq:optimization_problem_smoothed} decomposes the smoothed objective function as a sum of a strongly convex loss, $\Loss$, and the penalties.
Moreover, the least-squares loss, $\Loss(\bbeta) = \frac12\|\X\bbeta -\y\|_2^2$, can be re-written as a function of $\X\bbeta$ by $l(\z) \equiv \frac12\|\z - \y\|^2$, where $\z=\X\bbeta$. Therefore, we can equivalently express the smoothed objective function as
\begin{align}
    f_{\mu}(\bbeta) &= \Loss (\bbeta) + \Penalties_{\mu}(\bbeta) \nonumber\\
                    &= l(\X\bbeta) + \Penalties_{\mu}(\bbeta), \nonumber
\end{align}
\FIRST{where $\Penalties_{\mu}$ represents all penalty terms of \eqref{eq:optimization_problem_smoothed}.}
Our aim is to compute the duality gap to obtain an upper bound estimation of the distance to the optimum.
At any step $k$ of the algorithm, given the current primal $\bbeta^k$ and the dual $\bsigma(\bbeta^k)\equiv\nabla \Loss(\X\bbeta^k)$ variables~\cite{borwein2006convex}, we can compute the duality gap using the Fenchel duality rules~\cite{Mairal2010}.
This requires computing the Fenchel conjugates, $l^*$ and $\Penalties_\mu^*$, of $l$ and $\Penalties_{\mu}$, respectively.
While the expression of $l^*$ is straightforward, to the best of our knowledge, there is no explicit expression for $\Penalties_{\mu}^*$ when using a complex penalty such as TV or group Lasso.
Therefore, as an important theoretical contribution of this paper, we provide the expression for $\Penalties_{\mu}^*$ in order to compute an approximation of the duality gap that maintains its properties (\eqref{eq:gap_prop}).

\begin{theorem}[Duality gap for the smooth problem] \label{thm:gap_smooth}
    The following estimation of the duality gap satisfies \eqref{eq:gap_prop} \FIRST{, for any iterate $\bbeta^k$}:
    \begin{equation}\label{eq:gap_mu}
        \textsc{Gap}_{\mu}(\bbeta^k) \equiv f_\mu(\bbeta^k) +l^*(\bsigma(\bbeta^k)) + \Penalties_{\mu,k}^*(-\X^{\FIRST{\T}}\bsigma(\bbeta^k)),
    \end{equation}
    %\comment{$\Penalties_{\mu,k}$ is undefined (with subscript $k$) here. Why do we introduce the subscript $k$? I fail to wee why it is necessary.}
    with dual variable
    \begin{equation}
        \bsigma(\bbeta^k)\equiv \nabla l(\X\bbeta^k)= \X\bbeta^{\FIRST{k}} - \y,
    \end{equation}
    and the Fenchel conjugates
    \begin{align}\label{eq:gap_Fenchel_conjugates}
        l^*(\z) =& \frac{1}{2} \|\z\|_2^2 + \langle \z,\y\rangle  \nonumber\\
        \Penalties_{\mu,k}^*(\z) &\equiv
        \frac{1}{2\ltwoCoeff}\sum_{j=1}^\nDim \left(\bigg[\Big|z_j - \structCoeff \big(\A^\T \balpha_{\mu}^*(\bbeta^k)\big)_j\Big| - \loneCoeff\bigg]_+^2\right)  \nonumber\\
        &+ \frac{\structCoeff\mu}{2}\big\|\balpha_{\mu}^*(\bbeta^k)\big\|_2^2,
    \end{align}
    where $[\,\cdot\,]_+ = \max(0, \cdot\,)$.
\end{theorem}

The proof of this theorem can be found in \suppRef{supp:gap:proof:smooth}.
Note that there we also provide the expression and proof of the Fenchel conjugate for the non-smoothed problem, \ie, using $\Penalties$ instead of $\Penalties_{\mu}$.

The expression in \eqref{eq:gap_mu} of the duality gap of the smooth problem combined with the inequality in \eqref{eq:nesterov_inequality} provides an estimation of the distance to the minimum of the original non-smoothed problem. The sought distance is decreased geometrically by a factor $\tau\FIRST{\in(0, 1)}$ at the end of each continuation, and the decreased value defines the precision that should be reached by the next iteration (Line~\ref{algo:conesta:update_eps} of Algorithm~\ref{algo:conesta}).
%The particular value of $\tau$ appears to be of minor importance (according to our experimental results (data not shown)), and setting it to 0.5 is therefore assumed to work well in most situations.
Thus, the algorithm dynamically generates a sequence of decreasing precisions, $\varepsilon^i$. Such a scheme ensures the convergence (see \secref{sec:conesta:convergence}) towards a globally desired final precision, $\varepsilon$, which is the only parameter that the user needs to provide.

%%----------------------------------------------------------------------------%%
\subsection{Determining the optimal smoothing parameter} \label{sec:optimal_mu}
%%----------------------------------------------------------------------------%%

Given the current precision, $\varepsilon^i$, we need to compute a smoothing parameter $\mu_{opt}(\varepsilon^i)$ (Line~\ref{algo:conesta:update_mu} in Algorithm~\ref{algo:conesta}) that minimizes the number of FISTA iterations required to achieve such a precision when minimizing \eqref{eq:optimization_problem} via \eqref{eq:optimization_problem_smoothed} (\ie, such that $f(\bbeta^{k}) - f(\bbeta^*) < \varepsilon^i$). We have the following theorem giving the expression of the optimal smoothing parameter, for which a proof is provided in the \suppRef{supp:optimal_mu}.
\begin{theorem}[Optimal smoothing parameter, $\mu$] \label{thm:optimal_mu}
    For any given $\varepsilon>0$, selecting the smoothing parameter as
    \begin{equation} \label{eq:mu_opt}
        \mu_{opt}(\varepsilon) = \frac{-\structCoeff M \|\A\|_2^2 + \sqrt{(\structCoeff M \|\A\|_2^2)^2 + M L(\nabla(g))\|\A\|_2^2\varepsilon}}{ M L(\nabla(g))},
    \end{equation}
    minimizes the worst case bound on the number of iterations required to achieve the precision $f(\bbeta^k) - f(\bbeta^*) < \varepsilon$.
\end{theorem}
Note that $M=\nDim/{2}$ (\eqref{eq:nesterov_inequality}) and the Lipschitz constant of the gradient of $g$ as defined in \eqref{eq:optimization_problem_smoothed} is $L(\nabla(g)) = \lambda_{\max} (\X^\T \X) + \lambda$, where $\lambda_{\max}(\X^\T \X)$ is the largest eigenvalue of $\X^\T\X$.

%%----------------------------------------------------------------------------%%
\subsection{Convergence analysis of CONESTA} \label{sec:conesta:convergence}
%%----------------------------------------------------------------------------%%

%We call the resulting algorithm CONESTA.
The user only has to provide the global\FIRST{ly} prescribed precision $\varepsilon$, which will be guaranteed by the duality gap.
Other parameters are related to the problem to be minimized (\ie $g$, $\structCoeff$, $s$, $\loneCoeff$, $h$) and the encoding of the data structure $\A$.
%Finally, as mentioned above, $\tau$ can be fixed to $0.5$.
\FIRST{Finally, the value of $\tau$ was set to $0.5$.
Indeed, experiments shown in \suppRef{supp:sec:data:mri:tau} have demonstrated that values of $0.5$ or $0.2$ led to similar and increased speeds compared to larger values, such as $0.8$.
}

\begin{algorithm}
  \caption{CONESTA\big($\varepsilon$, $\A$, $g$, $s$, $h$, $\structCoeff$, $\loneCoeff$, $\tau=0.5$\big)}\label{algo:conesta}
  \begin{algorithmic}[1]
      \State Initialize $\bbeta^0 \in \setR^{\nDim}$
      \State $\varepsilon^0  = \tau \cdot \textsc{Gap}_{\mu=10^{-8}}(\bbeta^0)$ \label{algo:conesta:eps0}
      \State $\mu^0= \mu_{opt}\left(\varepsilon^0 \right)$ \label{algo:conesta:mu0}
      \Repeat
	\State $\varepsilon_\mu^i =\varepsilon^i-\mu^i \structCoeff M$ \label{algo:conesta:eps_mu}
	%\State $\bbeta_\mu^{i+1}= \textsc{FISTA}(\bbeta_\mu^i,\mu^i,\varepsilon_\mu^i, \ldots)$
        \State $\bbeta^{i+1} =$ \textsc{Fista}($\bbeta^{i}$, $\varepsilon_\mu^i$,  $\mu^i$, $\A$, $g$, $s_{\mu^i}$, $h$, $\structCoeff$, $\loneCoeff$)\label{algo:conesta:fista}
	\State $\varepsilon^i = \textsc{Gap}_{\mu=\mu^i} (\bbeta^{i+1}) + \mu^i \structCoeff M$ \label{algo:conesta:eps}
        %\STATE $N_{\mathrm{inner}} \leftarrow N_{\mathrm{inner}} + N_{\mathrm{f\mbox{}ista}}$
        \State $\varepsilon^{i+1} = \tau \cdot \varepsilon^i$\label{algo:conesta:update_eps}
        \State $\mu^{i+1} = \mu_{opt}\big(\varepsilon^{i+1}  \big)$ \label{algo:conesta:update_mu}
      \Until{$\varepsilon^i \leq \varepsilon$}\label{algo:conesta:stop}
      \State \textbf{return} $\bbeta^{i+1}$
  \end{algorithmic}
\end{algorithm}

CONESTA can be understood as a smooth touchdown procedure that uses the duality gap to probe the distance to the ground (global optimum) in order to dynamically adapt its speed (the smoothing).
Indeed, each continuation step of CONESTA (\algref{algo:conesta}) probes (Line~\ref{algo:conesta:eps}) an upper bound $\varepsilon^i$ of the current distance to the optimum ($f(\bbeta^i) - f(\bbeta^*)$) using the duality gap.
Then, Line~\ref{algo:conesta:update_eps} computes the next precision to be reached, $\varepsilon^{i+1}$, decreasing $\varepsilon^i$ by a factor $\tau \in (0, 1)$.
Line~\ref{algo:conesta:update_mu} derives the optimal smoothing parameter, $\mu^{i+1}$, required to reach this precision as fast as possible.
Finally, Line~\ref{algo:conesta:eps_mu} transforms back the precision with respect to the original problem into a precision for the smoothed problem, $\varepsilon_\mu^i$, using the inequality in \eqref{eq:nesterov_inequality}.
Therefore, at the next iteration, FISTA (Line~\ref{algo:conesta:fista}) will decrease $f_\mu^i$ until the error reaches $\varepsilon_\mu^i$.
Thanks to Line~\ref{algo:conesta:eps_mu}, this implies that the true error (toward the non-smoothed problem) will be smaller than $\varepsilon^i$.
The resulting weight vector, $\bbeta^{i+1}$, will be the initial value for the next continuation step using updated parameters.
Note that we use the duality gap for the smoothed problem, $\textsc{Gap}_{\mu=\mu^i}$ (and $\varepsilon_\mu^i$), and transform it back and forth using \eqref{eq:nesterov_inequality} to obtain the duality gap for the non-smooth problem, $\textsc{Gap}$ (and $\varepsilon^i$).
We do this because the gap on Line~\ref{algo:conesta:eps} has already been computed at the last iteration of the FISTA loop (Line~\ref{algo:conesta:fista}), since it was used in the stopping criterion.
Moreover, $\textsc{Gap}_{\mu}$ converges to zero for any fixed $\mu$ unlike $\textsc{Gap}$.
          
The initialization (Line~\ref{algo:conesta:eps0}) is a particular case where we use $\textsc{Gap}_{\mu}$ with a negligible smoothing value of \eg $\mu=10^{-8}$. We then derive the initial smoothing parameter on Line~\ref{algo:conesta:mu0}.
Therefore, if we start close to the solution the algorithm will automatically pick a small smoothing parameter, which makes CONESTA an excellent candidate for warm-restart.

%The following theorem presents the main steps of CONESTA.
The following theorem ensures the convergence and convergence speed of CONESTA.
\begin{theorem}[Convergence of CONESTA] \label{thm:convergence}
    Let $\left(\mu^i\right)_{i=0}^{\infty}$ and
    $\left( \varepsilon^i \right)_{i=0}^{\infty}$ be defined recursively by CONESTA (\algref{algo:conesta}). Then, we have that
    \begin{enumerate}[(i)]
       \item $\displaystyle{\lim_{i\rightarrow\infty}\varepsilon^i = 0}$,\quad and
       \item $f(\bbeta^i) \xrightarrow{i \rightarrow \infty} f(\bbeta^*)$.
       \item \SCND{Convergence rate of CONESTA with fixed smoothing (without continuation):
        For any given desired precision $\varepsilon >0$, using a fixed smoothing (line~\ref{algo:conesta:fista} of Algorithm~\ref{algo:conesta}) with an optimal value of $\mu$, equal to  $\mu_{opt}(\varepsilon)$, if the number of iterations $k$ is larger than}
        $$
            \frac{\sqrt{8\|\A\|_2^2M\structCoeff^2  \|\bbeta^0 - \bbeta^*\|^2_2 }}{\varepsilon} + \frac{\sqrt{2L(\nabla(g)) \|\bbeta^0 - \bbeta^*\|^2_2}}{\sqrt{\varepsilon}}.
        $$
        \SCND{then the obtained $\bbeta^k$ satisfies $f(\bbeta^k) - f(\bbeta^*) < \varepsilon$.}
       %\item \SCND{Convergence rate of CONESTA (with continuation) with the assumption of uniqueness of the minimum $\beta^*$. For any given desired precision $\varepsilon >0$ such that $f(\bbeta^i) - f(\bbeta^*) < \varepsilon$,  where the $\bbeta^i$'s are the sequence of solutions obtained from item (iii), then the smallest needed total sum of all the inner FISTA iterations is upper-bounded by}
	\item \SCND{Convergence rate of CONESTA (with continuation), assuming uniqueness of the minimum ($\beta^*$): For any given desired precision $\varepsilon >0$, if the total sum of all the inner FISTA iterations is larger than}
	%the smallest needed total sum of all the inner FISTA iterations is upper-bounded by}
        $$
            C/{\varepsilon},
        $$
        \SCND{where $C > 0$ is a constant, then the obtained solution (obtained from (iii)), \ie $\bbeta^i$, satisfies $f(\bbeta^i) - f(\bbeta^*) < \varepsilon$.}
    \end{enumerate}
\end{theorem}

\FIRST{The proof is provided} \SCND{Sec.~\ref{supp:thm:convergence:proof}} \FIRST{in the supplementary.}
\FIRST{Claim (i):} \SCND{Sec.~\ref{supp:thm:convergence:proof:i}} \FIRST{demonstrates that the sequence of decreased precisions $\varepsilon^i$ converges toward any prescribed precision.}
\FIRST{Claim (ii):} \SCND{Sec.~\ref{supp:thm:convergence:proof:ii}} \FIRST{demonstrates, at each step of the sequence, the solutions of the smoothed problem converge toward the solution of the non-smoothed problem.}
\FIRST{Claim (iii):} \SCND{Sec.~\ref{supp:thm:convergence:proof:iii}} \FIRST{demonstrates the number of iterations required to converge toward $\bbeta^*$ using the auxiliary smoothed problem (without continuation) with a fixed and optimal smoothing value.}
\FIRST{Finally, claim (iv),} \SCND{Sec.~\ref{supp:thm:convergence:proof:iv}} \FIRST{provides the convergence rate with respect to the total number of iterations.}

The continuation technique improves the convergence rate compared to the simple smoothing using a single value of $\mu$. Indeed, it has been demonstrated in \cite{teb2} (see also \cite{Chen2012}) that the convergence rate obtained with a single value of $\mu$, even optimized, is $\bigO{1/\varepsilon} + \bigO{1/{\sqrt{\varepsilon}}}$. However, the CONESTA algorithm achieves $\bigO{1/\varepsilon}$ for simply \FIRST{(non-strongly)} convex functions.

In \theoremref{thm:convergence} we need the uniqueness hypothesis in (iv) to obtain the boundedness of the FISTA iterations. The authors in \cite{Chambolle2015} have proved this boundedness property for a slightly modified version of the FISTA algorithm and general convex function. Nevertheless, we could not use these results because it should be adapted to the smoothed version, which has not been done yet to the best of our knowledge. This would require establishment of the boundedness of the FISTA iterated uniformly with respect to $\mu$ as it converges to zero.

%\theoremref{thm:convergence} use only a weak hypothesis of the boundedness of the FISTA iteration and avoid strong convexity hypothesis for tow reasons:

%With strong convexity hypothesis we could not use results from \cite{Chambolle2015} that has proven the boundedness property for a slightly modified version of the FISTA algorithm because not adapted to the smoothed version. This would require to establish the boundedness of the FISTA iterated uniformly with respect to $\mu$ as it converges to zero, which has not been done yet to the best of our knowledge.

%Note that the authors in \cite{Chambolle2015} have proved this boundedness property for a slightly modified version of the FISTA algorithm. We do not use these results because it should be adapted to the smoothed version which has not been done yet to the best of our knowledge. This would require to establish the boundedness of the FISTA iterated uniformly with respect to $\mu$ as it converges to zero.

%A second argument to avoid the strong convexity hypothesis,
Moreover, we avoid the strong convexity hypothesis because if the objective function is strongly convex with strong convexity parameter of the $g$ function, then FISTA enjoys global linear convergence rate and especially ensures the convergence of $\bbeta^k$ to the minimum, $\bbeta^*$. In fact, we have that $\|\bbeta^k - \bbeta^*\|_2 < \sqrt{{2\sigma}/\varepsilon}$ as far as $f(\bbeta^k)-f(\bbeta^*) < \varepsilon$ (see Remark~3 in \cite{Chen2012}). In this case, not only is the hypothesis in (iv) naturally satisfied, but it is also a favorable case to apply a continuation technique since the estimations on the successive intermediate $\bbeta^k$ will ensure a faster convergence contrary to the simple convergence case in which we don't make use of this rate. Nevertheless, we did not include this case in this paper since it will engender the modification of the optimal smoothing value $\mu_{opt}$ and consequently all the other estimations due to the different convergence rate of FISTA. These questions will \FIRST{instead} be considered in future works.

%%%%%%%%%%%%%%%%%%%%%%%%%%%%%%%%%%%%%%%%%%%%%%%%%%%%%%%%%%%%%%%%%%%%%%%%%%%%%%%%
\section{Experiments} \label{sec:experiments}
%%%%%%%%%%%%%%%%%%%%%%%%%%%%%%%%%%%%%%%%%%%%%%%%%%%%%%%%%%%%%%%%%%%%%%%%%%%%%%%%

In this section we compare CONESTA to the state-of-the-art algorithms mentioned above (see \suppRef{supp:sec:state-of-the-art-solvers:details} for details), \ie, ADMM, EGM, Inexact FISTA and FISTA with fixed $\mu$. We will use these algorithms to solve the problem on both simulated and high-dimensional structural neuroimaging data.

We used FISTA with fixed $\mu$ using two values of $\mu$, chosen as follows: (i) Chen's $\mu$ where $\mu = \varepsilon/(2\structCoeff M)$ as was used in~\cite{Chen2012} and (ii) Large $\mu = (\text{Chen's}~\mu)^{1/2}$. The first proposal for $\mu$ ensures that we reach the desired precision, although convergence may be slow. The second proposal has a value of $\mu$ that may not guarantee we reach the desired precision before convergence.

%%----------------------------------------------------------------------------%%
\subsection{Simulated 1D data set where the minimum of $f$ is known} \label{sec:data:simu}
%%----------------------------------------------------------------------------%%

We generated simulated data where we control the true minimizer, $\bbeta^*$, and the associated regularization parameters, $\loneCoeff$, $\ltwoCoeff$ and $\structCoeff$~\cite{Lofstedt_simulated_2015}.
%\subsubsection{Data set generation}
The experimental setup for the simulated 1D data set was inspired by that of~\cite{Bach_etal_2011} and is shown in \tabref{tab:simulation_design}. This setup is a designed experiment with multiple small- and medium-scale data sets having combinations of low, medium and high degrees of correlation between variables, low, medium and high levels of sparsity, and low, medium and high signal-to-noise ratios.

\begin{table}[!h]
    \begin{center}
        \caption{The experimental setup for the simulation study. The
                 parameters varied were: the size of the data set, $(n, p)$, the
                 correlation between variables, the sparsity of the data, and
                 the signal-to-noise ratio. Five runs were performed for each
                 combination of settings in order to assess the variability of the different algorithms. The medium level for sparsity (marked with an asterisk, `*') was only performed for the low and medium sizes.}
        \begin{tabular}{llcccc}
        \toprule
        Level  &\phantom{}& Size ($n\times p$) & Correlation & Sparsity    & Signal-to-noise ratio \\
        \cmidrule{1-1} \cmidrule{3-6}
        Low    && $200\times200$     & Low         & 50~\%       & 0.5 \\
        Medium && $632\times1514$    & Medium      & 72.5~\%$^*$ & 1.0 \\
        High   && $2000\times10000$  & High        & 95~\%       & 5.0 \\
        \bottomrule
        \end{tabular}
        \label{tab:simulation_design}
    \end{center}
\end{table}

A candidate version of the predictors, denoted $\X_0$, was drawn from a multivariate Gaussian distribution $\mathcal{N}(\mathbf{1},\Sigma)$, where $\Sigma$ was generated according to the constant correlation model described in~\cite{Hardin_etal_2013}, where $\diag(\Sigma)=\mathbf{1}$ and off-diagonal elements are all equal to $\rho$ which is drawn from $\mathcal{N}(0,d_c/\sqrt{n})$.
%The $\rho$ is drawn from a distribution that approximately follows the Gaussian distribution $\mathcal{N}(0,d_c/\sqrt{n})$.
Here, $d_c$ controls the dispersion of the correlation and takes values of $d_c=1, 4.5$ and $8$ for low, middle and high correlation cases.
%Concerning the weight vector $\bbeta^*$, a sparsity parameter fixes the percent of null weights within it.
\FIRST{A sparsity parameter (Tab. \ref{tab:simulation_design}) fixes the percent of null weights within weight vector $\bbeta^*$.}
The remaining weights that contribute to the prediction were drawn from $\mathcal{U}(0,1)$ and sorted in ascending order.
The residuals, $\mathbf{e}$, were drawn from a Gaussian distribution, $\mathcal{N}(1,1)$, and then scaled to unit $\ell_2$-norm.
Given a candidate data matrix, $\X_0$, the true minimizer $\bbeta^*$, the true residual vector $\e$, and regularization constants, $\loneCoeff$, $\ltwoCoeff$ and $\structCoeff$, that were arbitrarily set to $\loneCoeff=0.618$, $\ltwoCoeff=1-\loneCoeff$ and $\structCoeff=1.618$, we generated the simulated data set $(\X, \y)$ such that $\bbeta^*$ is the solution that minimizes \eqref{eq:optimization_problem}, with $s(\bbeta^*)=\TV(\bbeta^*)$.
%The data matrix, $\X$, is obtained by finding scaling factors of each column of the candidate data matrix, $\X_0$.
\FIRST{The final data matrix, $\X$, is obtained by applying a scaling factor to each column of the candidate data matrix, $\X_0$.}
All details of the scaling procedure to produce a data set with a known exact solution are described in the ancillary paper~\cite{Lofstedt_simulated_2015}.
Once $\X$ is found, the $\y$ vector is computed as $\y = \X\bbeta^* - \mathbf{e}$.
%The main benefit of generating the data in this way is that we know everything about it. In particular, we know the true minimizer, $\bbeta^*$, and we know the regularization parameters $\loneCoeff$, $\ltwoCoeff$ and $\structCoeff$.
%There is no need to use \eg cross-validation to find any parameters, and we can immediately determine whether the $\bbeta^k$ we obtain is close to the true $\bbeta^*$ or not.
For each combination of settings, we generated data knowing the exact solution $\bbeta^*$ of the given minimization problem.
Thus, we could compute the error $f(\bbeta^k) - f(\bbeta^*)$ for any current solution $\bbeta^k$.
Five runs were performed for each combination of the settings, resulting in a total of $405$ simulation runs.

\subsubsection{Result of the comparison}\label{sec:data:simu:results}

We applied the minimization algorithms to simulated data that were generated
using the settings shown in \tabref{tab:simulation_design}. For each of the
settings of the designed experiment, we measured the number of iterations and
the time (in seconds) required to reach a certain precision level,
$\varepsilon$, for each of the tested minimization algorithms.
For each precision level, ranging from $1$ to $10^{-6}$, for each algorithm
and for each data set, we \textit{ranked} each algorithm according
to the time they required to reach a given precision. Then, we averaged
the ranks across data sets (see \tabref{tab:algo_rank_friedman_test}) and we tested (Friedman test ~\cite{friedman_comparison_1940}) the significance of the
difference in ranks between algorithms.

\begin{table}[!ht]
\begin{center}
\caption{Average rank of the convergence speed of the algorithms to reach
    precisions ($f(\bbeta^k) - f(\bbeta^*)$) ranging from 1 to $10^{-6}$.
    We have here reported whether the average rank of a given algorithm was
    significantly larger $\bm{>}$ (slower) or significantly smaller $\bm{<}$
    (faster) than CONESTA (a missing `$\bm{>}$' or `$\bm{<}$' means
    non-significant).
    \FIRST{P-values were calculated with a post hoc analysis of the Friedman test corrected
    for multiple comparisons. Note that all reported significant differences had a corrected p-value of $10^{-3}$ or smaller.}
    %We performed a post hoc analysis using the Friedman
    %test. All significant results reported were $p < 10^{-3}$ or smaller, corrected
    %for multiple comparisons.
    For a given data set, all algorithms were
    evaluated with a limited upper execution time. Thus, some high precisions
    (\eg, $10^{-5}, 10^{-6}$) were not always reached within the limited time.
    For FISTA with a large $\mu$, the high precisions may in fact not be reachable at all. In those situations, the execution time was
    set to $+\infty$.
    }

\setlength{\tabcolsep}{3pt} % General space between cols (6pt standard)
    
\begin{tabular}{@{}lrr@{}rrr@{}rr@{ }r@{}rr@{ }r@{}rr@{ }r@{}rr@{ }r@{}rr@{ }r@{}r@{ }}
\toprule
&& \multicolumn{20}{c}{Average rank of the time to reach a given precision}\\
\cmidrule{3-22}
Algorithm             
&&\multicolumn{2}{c}{1}
&&\multicolumn{2}{c}{$10^{-1}$}
&&\multicolumn{2}{c}{$10^{-2}$}
&&\multicolumn{2}{c}{$10^{-3}$}
&&\multicolumn{2}{c}{$10^{-4}$}
&&\multicolumn{2}{c}{$10^{-5}$}
&&\multicolumn{2}{c}{$10^{-6}$}\\
%\midrule
\cmidrule{1-1} \cmidrule{3-4} \cmidrule{6-7} 
\cmidrule{9-10} \cmidrule{12-13} \cmidrule{15-16} 
\cmidrule{18-19} \cmidrule{21-22}
CONESTA               &&3.3        &  --  &&2.7&--      &&2.2& --     && {\bf 1.6}&--&&{\bf 1.3}&--&&{\bf 1.0}&--&&{\bf 1.0}&--\\
ADMM                  &&2.9        &      &&{\bf 2.1}&  &&{\bf 1.9}&  && 1.8&        &&1.7&$\bm{>}$&&1.8&$\bm{>}$&&1.5&$\bm{>}$\\
EGM         &&{\bf 1.8}&$\bm{<}$&&2.2&        &&2.0      &  && 2.3&$\bm{>}$&&2.3&$\bm{>}$&&2.2&$\bm{>}$&&1.7&$\bm{>}$\\
FISTA large $\mu$    &&4.7&$\bm{>}$      &&6.0&$\bm{>}$&&4.6&$\bm{>}$&& 3.4&$\bm{>}$&&2.7&$\bm{>}$&&2.2&$\bm{>}$&&1.7&$\bm{>}$\\
%FISTA large $\mu$     &&2.9        &      &&3.2&        &&3.7&$\bm{>}$&& 3.4&$\bm{>}$&&2.7&$\bm{>}$&&2.2&$\bm{>}$&&1.7&$\bm{>}$\\
FISTA Chen's $\mu$    &&6.2&$\bm{>}$      &&5.0&$\bm{>}$&&4.4&$\bm{>}$&& 3.4&$\bm{>}$&&2.7&$\bm{>}$&&2.2&$\bm{>}$&&1.7&$\bm{>}$\\
Inexact FISTA         &&6.2&$\bm{>}$      &&5.4&$\bm{>}$&&4.3&$\bm{>}$&& 3.3&$\bm{>}$&&2.7&$\bm{>}$&&2.2&$\bm{>}$&&1.7&$\bm{>}$\\
\bottomrule
\end{tabular}

\label{tab:algo_rank_friedman_test}
\end{center}
\end{table}

\tabref{tab:algo_rank_friedman_test} indicates that EGM is the fastest algorithm for the lowest (\ie, largest) precision, $\varepsilon=1$, and it significantly outperformed CONESTA. ADMM was the second fastest algorithm, but it was not significantly faster than CONESTA. CONESTA was the third fastest and significantly outperformed the Inexact FISTA and FISTA with either $\mu$.

The average convergence rank of CONESTA improved for the higher precision
levels. However, it remained the third fastest algorithm for levels
\(\varepsilon=10^{-1}\) and \(\varepsilon=10^{-2}\), after ADMM and the
EGM, but was now no longer significantly slower than either algorithm.
However, superiority over all FISTA with Nesterov's smoothing and
Inexact FISTA remained and became significant.
For higher (smaller than $10^{-3}$) precisions, CONESTA outperformed
all other solvers. Its superiority was significant for all
precisions, except for the comparison with ADMM at $\varepsilon=10^{-3}$.

\begin{figure}[H]
\captionsetup[subfloat]{farskip=2pt,captionskip=1pt}
  \centering
  \subfloat[Simulated data sets of size $200 \times 200$.]{\label{fig:edge-a}\includegraphics[width=0.5\textwidth]{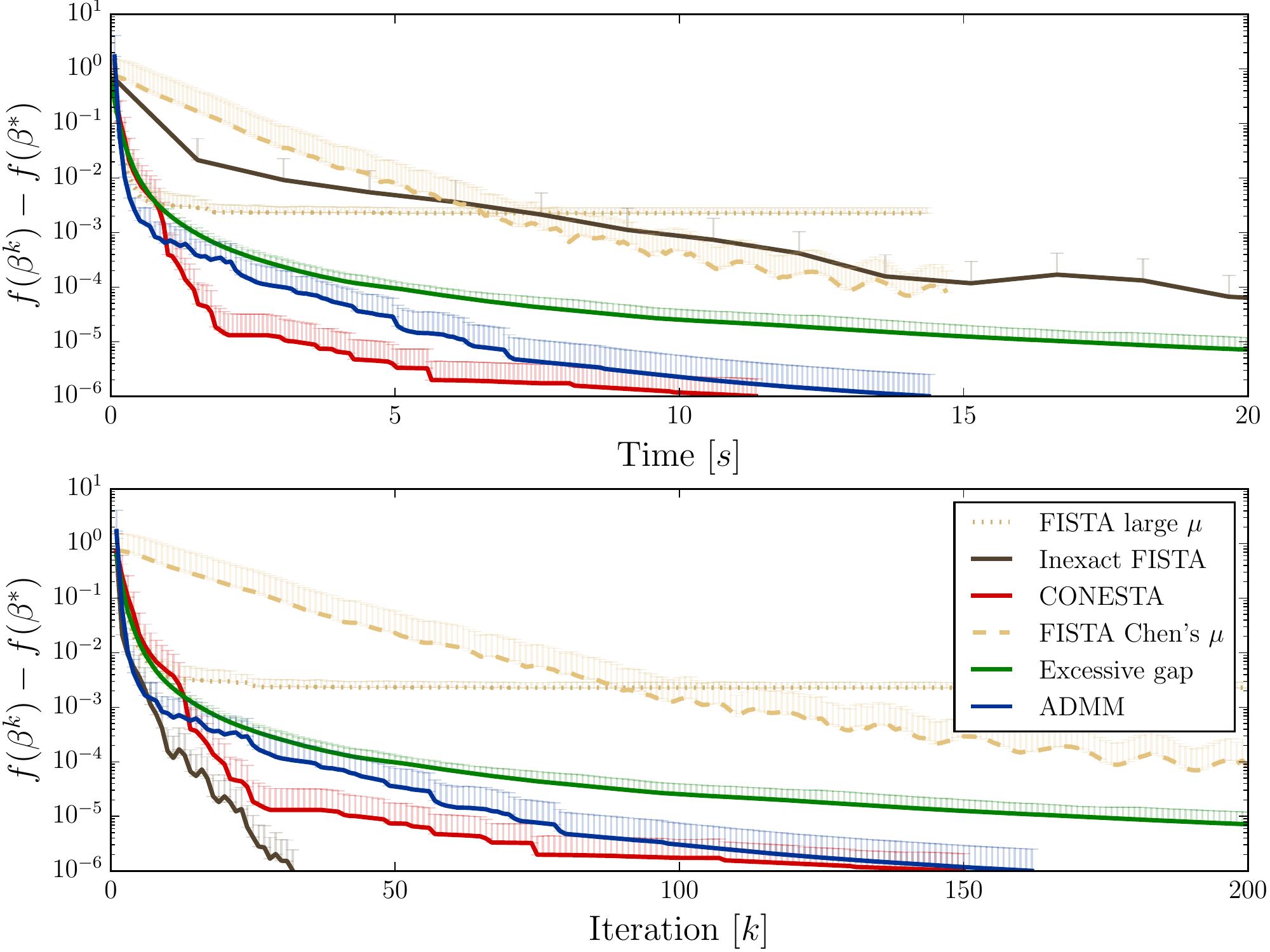}}\\
  %\hspace{5pt}
  \subfloat[Simulated data sets of size $632 \times 1514$.]{\label{fig:edge-a}\includegraphics[width=0.5\textwidth]{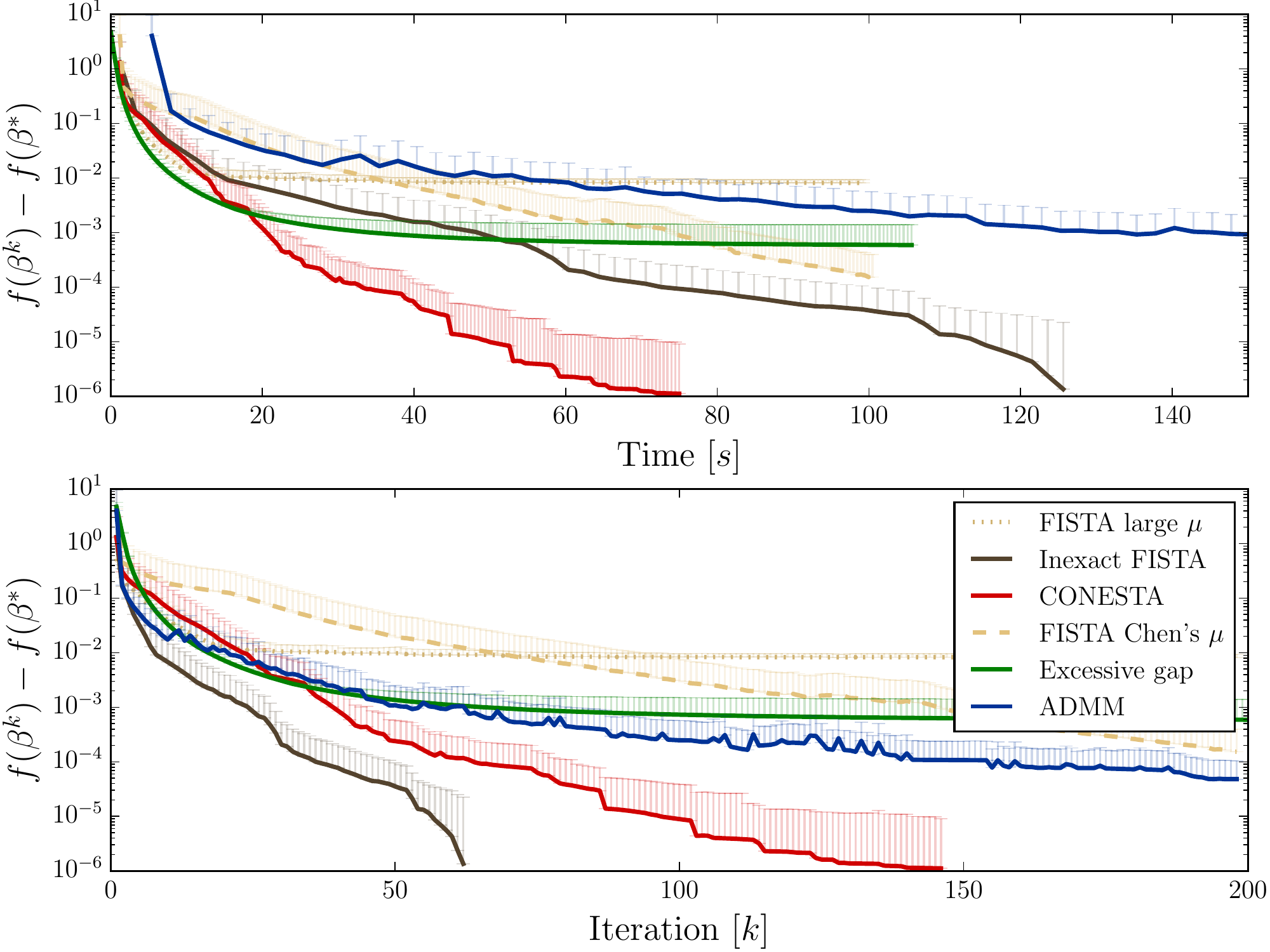}}\\
  %\hspace{5pt}
  \subfloat[Simulated data sets of size $2000 \times 1000$.]{\label{fig:edge-a}\includegraphics[width=0.5\textwidth]{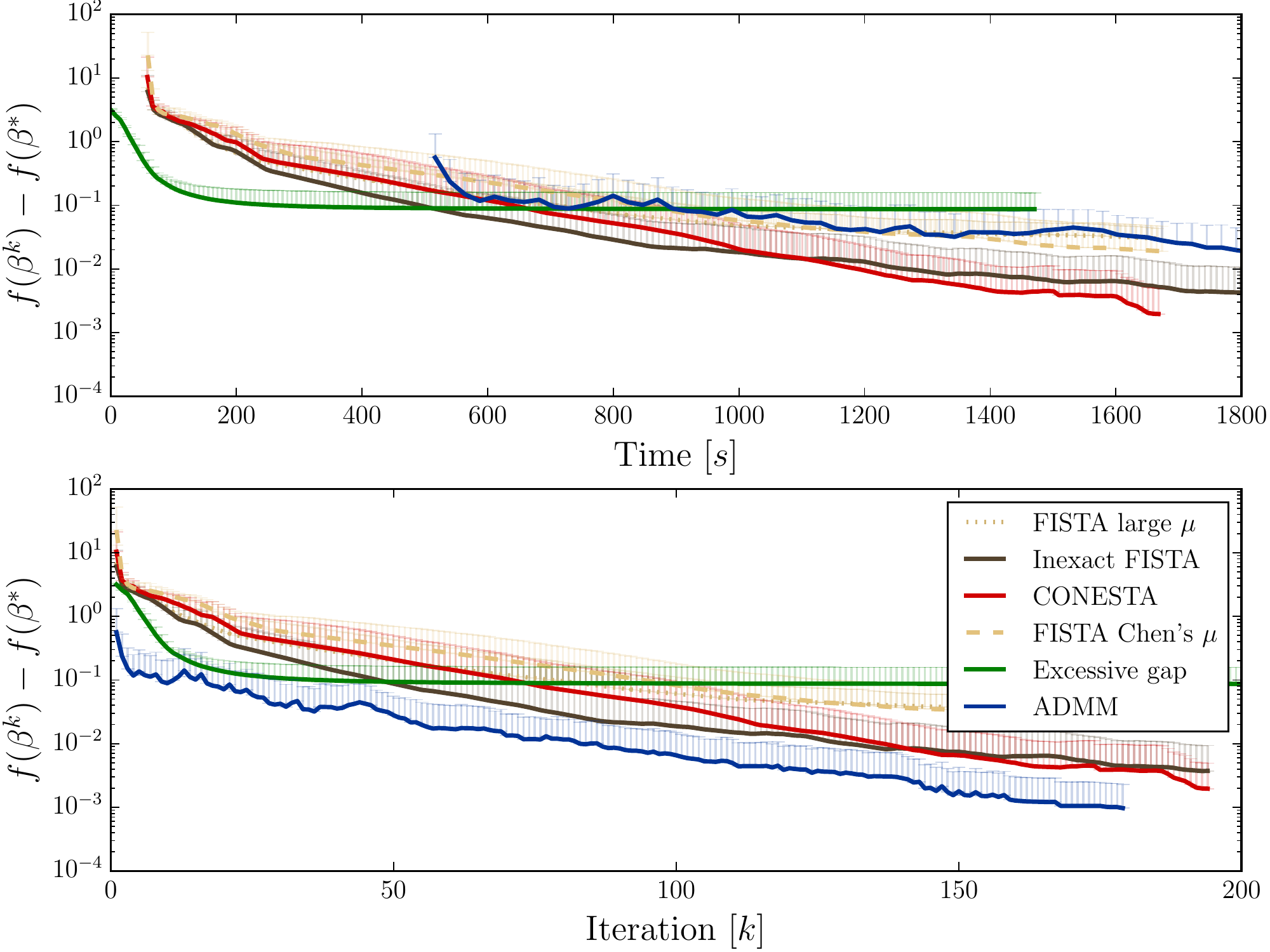}}
  \caption{For datasets of different sizes: (a) $200 \times 200$, (b) $632 \times 1514$, (c) $2000 \times 1000$, 
	the y-axis shows the median error (+MAD), over 
	all experimental settings, as a function of the computational time (top plot) and
	the number of iterations (bottom plot).}
  \label{fig:algos-convergence_simulated}
\end{figure}

\figref{fig:algos-convergence_simulated} presents the median error (\ie, the median over $f(\bbeta^k) - f(\bbeta^*)$, and the median absolute deviation, MAD), over all experimental settings (\tabref{tab:simulation_design}) of correlation, sparsity and signal-to-noise ratio, as a function of the time (top plot) and the number of iterations (bottom plot).
Results are shown separately for each different sizes of data set: (a) $200 \times 200$, (b) $632 \times 1514$ and (c) $2000 \times 10000$.
%For each data set size, we computed the median error (\ie, the median over $f(\bbeta^k) - f(\bbeta^*)$) obtained by each solver (and the median absolute deviation), over all experimental settings of correlation, sparsity and signal-to-noise ratio of the data.
%\figref{fig:algos-convergence_simulated} provides details on the solvers behaviors on data sets of different sizes, and the median errors are reported as functions of the computational time (top plot) and the numbers of iterations (bottom plot).
Note that we did not perform any particular code optimization in order to fairly measure the time difference between the algorithms.
Moreover, many solvers share large portions of code, especially FISTA, Inexact FISTA and CONESTA.
\FIRST{All solvers are provided by the ParsimonY Python library (see Appendix \secref{sec:parsimony}). The experiments were conducted on a Linux-based operating system (Ubuntu 16.04 LTS)} \SCND{using a single core of an Intel i7-5600U@2.6~GHz processor with 16~GB of RAM}.

%However, to avoid any suspicion of bias due to our implementation, \figref{fig:algos-convergence_simulated} presents the error as a function of the number of iterations.

The figure confirms the efficacy of EGM for the lowest precision, at least for the small and medium size data sets. Then, on all data sets, EGM appears to be slow to reach higher precisions.
ADMM appears to be particularly fast in the beginning where
it converges quickly to the middle precision levels, after which it begins to
level out and be outperformed by CONESTA. This ADMM behavior is well-known
and is discussed in \eg~\cite{Boyd_etal_2011}.
The Inexact FISTA performs impressively when only looking at the number of iterations.
However, if we consider the time that takes in account the cost of the inner loop for Inexact FISTA, it is almost always outperformed by CONESTA at the higher precisions ($\varepsilon<10^{-1}$), which correspond to tens of outer FISTA iterations and hundreds of approximation iterations.

CONESTA demonstrates a stable behavior: whatever the setting of the data set, it is always among the best solvers for low precisions and it outperforms others solvers for higher precisions ($\varepsilon \leq 10^{-3}$).

%%----------------------------------------------------------------------------%%
\subsection{Experiments on a structural MRI data set}\label{sec:data:mri}
%%----------------------------------------------------------------------------%%

We applied the solvers on a structural MRI data set of 199 subjects from the Alzheimer's Disease Neuroimaging Initiative (ADNI, \url{http://adni.loni.usc.edu/}) cohort.
We included 119 controls and 80 patients with mild cognitive impairment (MCI) that converted to AD within 800 days, see \suppRef{supp:sec:data:mri} for details.
Our goal was to predict the subjects' ADAS (AD Assessment Scale-Cognitive Subscale) score ($\y$), measured 800 days after brain image acquisition.
%As input ($\X$), we retained $286\,214$ voxels of gray matters extracted with SPM8 using DARTEL normalization~\cite{ashburner_unified_2005}.
\SCND{Images were segmented (SPM8) and spatially normalized into a template (voxels size: 1.5 mm isotropic) using DARTEL~\cite{ashburner_unified_2005}.
Gray matters (GM) probability maps were modulated with the Jacobian determinants of the nonlinear deformation field.
$286\,214$ voxels of GM were retained and concatenated to form the input data $\X$.}

\subsubsection{Relevance of the TV penalty in the context of neuroimaging}\label{sec:data:mri:relevance}
%Predictive Support Recovery with the TV-Elastic Net Penalty}

Before comparing convergence speed of the solvers, this section demonstrates the improvements obtained with structured sparsity ($\ell_1$ and TV) in the context of neuroimaging.
%Indeed, the ultimate purpose of CONESTA is to provide an efficient solver for prediction with structured sparsity on high-dimensional neuroimaging data.
We compared Lasso ($\ell_1$), Ridge regression ($\ell_2$), elastic net ($\ell_1+\ell_2$), with and without the TV penalty, using 5-fold cross-validation (5CV), in terms of (i) prediction performance (coefficient of determination, $R$-squared, $R^2$); and (ii) more importantly, the stability of the $\bbeta$ maps against variations of the learning samples. Indeed, clinicians expect that the identified neuroimaging predictive signature, \ie the non-\FIRST{zero} weights of the $\bbeta$ map, to be similar if other patients with similar clinical conditions would have been used.

We used two similarity measures to assess the stability of those $\bbeta$ maps across the re-sampling. First, we computed the mean correlation between pairs of $\bbeta$ maps obtained across the 5CV: pairwise correlations were transformed into $z$-scores using Fisher's $z$-transformation, averaged and finally transformed back into an average correlation, denoted $\overline{r}_{\bbeta}$. Second, we computed an inter-rater agreement measure: the Fleiss' \(\kappa\) statistics, to assess the agreement between supports (non-\FIRST{zero} weights of the $\bbeta$ maps) recovered by the different penalties across the 5CV.
The weights of the five $\bbeta$ maps were partitioned into three categories according to their signs. This led to negative, positive and out-of-support voxels. The Fleiss' kappa statistic, $\kappa_{\bbeta}$, was then computed for the five raters. As the 5CV folds share 60~\% of their training samples, no unbiased significance measure can be directly obtained from the $\overline{r}_{\bbeta}$ and $\kappa_{\bbeta}$ statistics. Thus, we used permutation testing to assess empirical $p$-values.
%The null hypothesis was simulated by $1\,000$ random permutations of the target variable, $\y$, within each permuted sample we executed the whole 5CV round. Then the statistics on the true data were compared with the ones obtained on the permuted ones.

\begin{table}[!h]
    \centering
    \caption{Predictive performances: $R$-squared ($R^2$) and stability of 
             $\bbeta$ maps measured as the average correlation between $\bbeta$ map pairs ($\overline{r}_{\bbeta}$, using Fisher's $z$-transformation) and the inter-rater measure ($\kappa_{\bbeta}$, Fleiss' \(\kappa\) statistic) of agreement between the supports recovered across the 5CV. Significance notations:
             ***:~$p\leq10^{-3}$, **:~$p\leq10^{-2}$, *:~$p\leq0.05$.}
             %Empirical $p$-values were assessed by $1\,000$ executions of the 5CV round on randomly permuted data.
%{\tiny
\setlength{\tabcolsep}{1pt} % General space between cols (6pt standard)
\renewcommand{\arraystretch}{1} % General space between rows (1 standard)

\begin{tabular}{@{}ll@{}lll@{}ll@{}l@{}}
\toprule
                    &  \multicolumn{2}{c}{Prediction}
                        & \phantom{a}
                            &  \multicolumn{4}{c}{Stability of $\bbeta$ maps}\\
                    \cmidrule{2-3}  \cmidrule{5-8}
Method              & $R^2$
                        &~~~$\Delta$
                            &
                    & $\overline{r_{\bbeta}}$
                        &~~~$\Delta$
                            & $\kappa_{\bbeta}$
                                &~~~$\Delta$\\
\midrule
$\ell_1$            & 0.50*
                        &\multirow{2}{*}{{\large\}}$+0.02$**}
                            &
                    & 0.58***
                        &\multirow{2}{*}{{\large\}}$+0.09$***}
                            & 0.29***
                                &\multirow{2}{*}{{\large\}}$+0.07$***}\\
$\ell_1+TV$         & 0.52**
                        &
                            &
                    & 0.67***
                        &
                            & 0.36***
                                & \\
$\ell_1+\ell_2$     & 0.53*
                        & \multirow{2}{*}{{\large\}}$-0.02$}
                            &
                    & 0.58***
                        & \multirow{2}{*}{{\large\}}$+0.07$***}
                            & 0.33***
                                & \multirow{2}{*}{{\large\}}$+0.02$***}\\
$\ell_1+\ell_2+TV$  & 0.51**
                        &
                            &
                    & 0.65***
                        &
                            & 0.35***
                                & \\
\bottomrule
\end{tabular}
%}
 \label{tab:results-mri-tv}
\end{table}

The predictive performances, displayed in \tabref{tab:results-mri-tv}, were not markedly improved with the TV penalty. However, using TV in conjunction with the $\ell_1$ penalty was found to significantly improve predictive performance as compared with the usage of $\ell_1$ alone.

\begin{figure}[!h]
    \centering
    \includegraphics[width=0.5\textwidth]{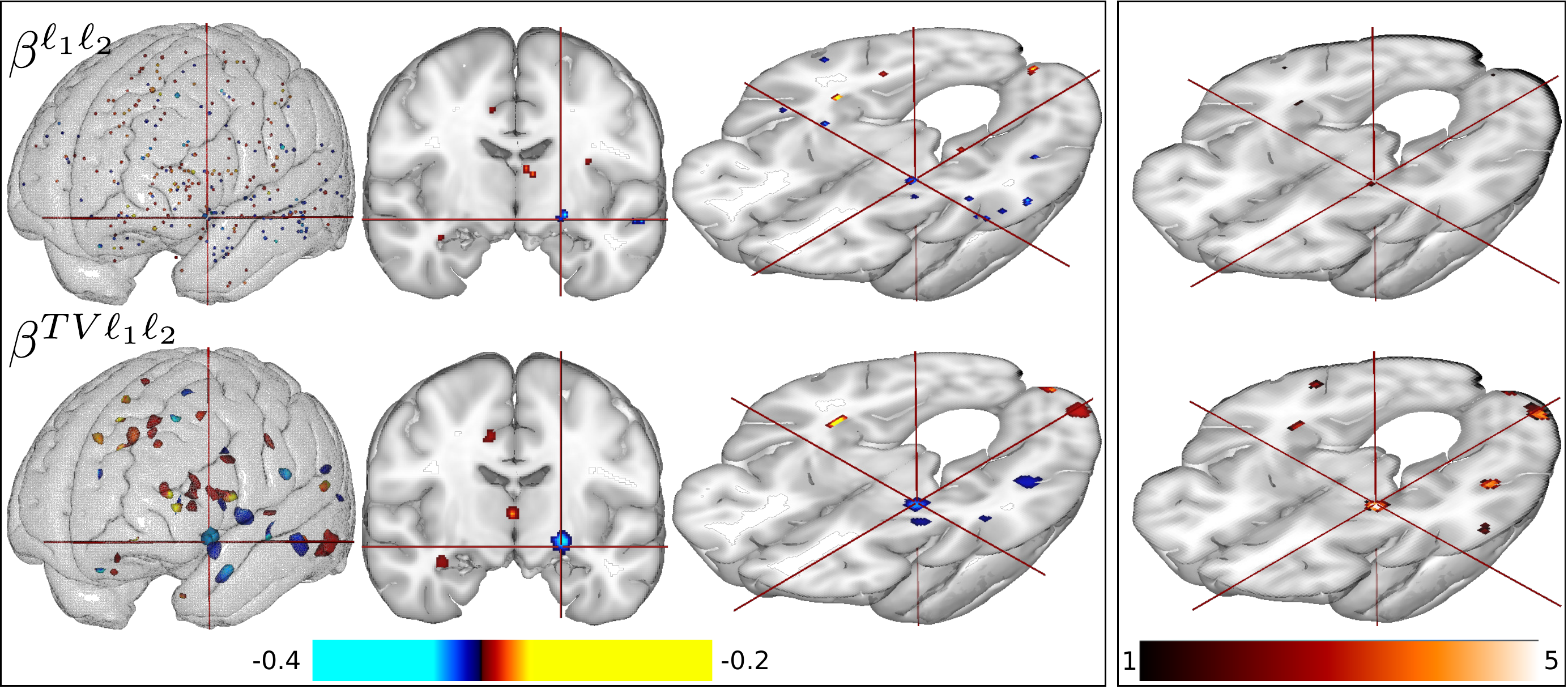}
    \caption{Left panel: weight map found with elastic net
             ($\bbeta^{\ell_1\ell_2}$) and with the elastic net and TV penalties ($\bbeta^{\ell_1\ell_2TV}$) found using CONESTA. Right panel: counts of the number of times each voxels were part of the support (non-\FIRST{zero} weights) across the 5CV.}
\label{fig:weights_map_cvcount}
\end{figure}

\figref{fig:weights_map_cvcount} demonstrates that the TV penalty provides a major breakthrough regarding support recovery of the predictive brain regions. Contrary to the elastic net penalty that highlights scattered and seemingly meaningless patterns, when using the elastic net \FIRST{penalties with} TV (\ie, $\ell_1+\ell_2+TV$) the parameter map is smooth and corresponds to brain regions known to be involved in AD~\cite{frisoni_clinical_2010}. The most important cluster of non-\FIRST{zero} weights, found in the left hippocampus and the ventricular enlargement, was identified. The right panel of \figref{fig:weights_map_cvcount} confirms that TV produces more stable supports than when using only elastic net. Without TV, some (few) voxels within the hippocampus were at most selected three times during the 5CV rounds. These results demonstrate the ineffectiveness of elastic net in the context of biomarker identification. Conversely, with the TV penalty, voxels within the hippocampus were always selected. The measures of the stability of $\bbeta$ maps presented in \tabref{tab:results-mri-tv} demonstrated that TV significantly improves the reproducibility of the parameter maps, leading to meaningful predictive signatures.

\subsubsection{Comparisons of the convergence speed of the minimization algorithms} \label{sec:structural_mri_data}

After having demonstrated the relevance of TV for extracting stable signatures, we revisit the comparison of the convergence speed of CONESTA to that of the state-of-the-art solvers. In this case: EGM, FISTA with two different fixed $\mu$ and Inexact FISTA.
Note that ADMM was excluded in this example. The version in \cite{Wahlberg_etal_2012} is made for 1D underlying data, and the authors did not adapt the method to the case of 2D or 3D underlying data.
%It would be
%possible to represent the anisotropic total variation in this form when the
%underlying data has more than one dimension, but that is obviously a different
%penalty and thus a different optimization problem.
Note also that without the particular form of the \(\A^\T\A\) matrix (it is tridiagonal in the \(\ell_1+\TV_{1D}\) case), ADMM would perform much slower, because it requires us to solve a linear system in each iteration. This is thus a major drawback of ADMM, that it cannot easily be adapted to arbitrary complex penalties.
%~\\[0.0em]

\begin{figure}[!h]
     \centering
     \includegraphics[width=0.5\textwidth]{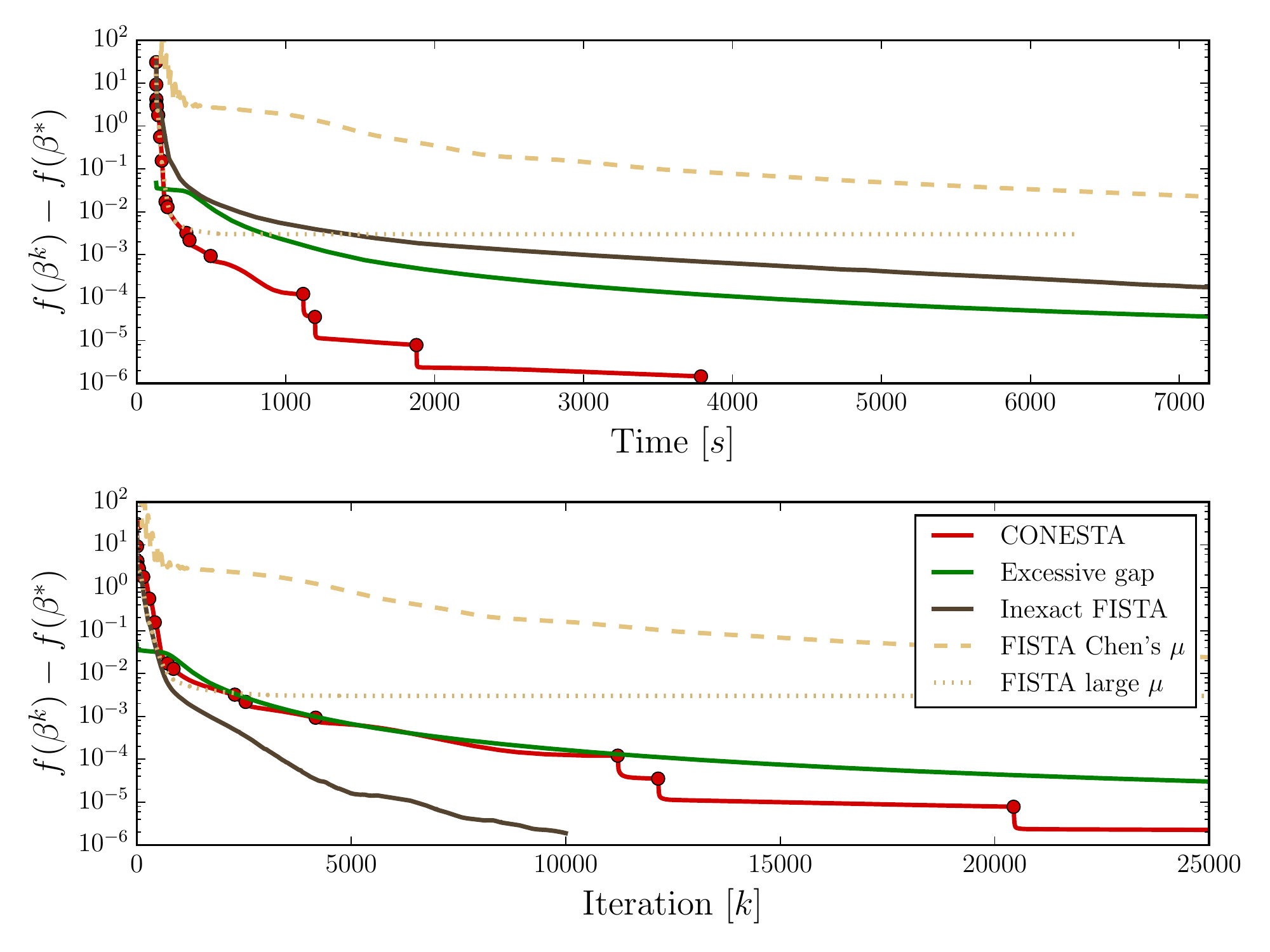}
     \caption{The error as a function of the computational time (top plot) and
              the number of iterations (bottom plot). The vertical axis is plotted
              on a logarithmic scale. Dots on the CONESTA curve indicate where
              the continuation steps take place, \ie where the dynamic selection of
              a new smoothing parameter happened.}
\label{fig:algos-convergence_mri}
\end{figure}

First, we fixed a desired precision, $\epsilon=10^{-6}$, that was used as the stopping criterion for all algorithms. It was also used to derive the smoothing parameter for FISTA with fixed $\mu$. We ran CONESTA and Inexact FISTA until they reached a precision of $10^{-7}$, evaluated with the duality gap. The smallest value of $f(\bbeta^k)$ was considered as the global minimum $f(\bbeta^*)$ used to compute the errors $f(\bbeta^k) - f(\bbeta^*)$ in \tabref{fig:algos-convergence_mri} and \tabref{tab:algos-convergence_mri}.

\begin{table}[!ht]
    \centering
    \caption{Execution time ratios of each state-of-the-art algorithm over the
             time required by CONESTA to reach the same precision.}
    \begin{tabular}{lrrrrrrr}
        \toprule
                         &  \multicolumn{7}{c}{Time ratio over CONESTA to reach a given precision}\\
        \cmidrule{2-8}
               Algorithm & $10^{1}$ & $10^{0}$ & $10^{-1}$ & $10^{-2}$ & $10^{-3}$ & $10^{-4}$ & $10^{-5}$ \\
        \midrule
%             CONESTA     &      1.0 &      1.0 &       1.0 &       1.0 &       1.0 &      1.0 &    1.0 \\
           EGM           &     0.97 &     0.39 &      0.24 &      1.05 &      1.08 &     1.74 &   3.64 \\
         Inexact FISTA   &     1.13 &     1.92 &      2.64 &      3.97 &      4.38 &     5.45 &   6.07 \\
      FISTA Chen's $\mu$ &      2.7 &    10.65 &      17.5 &     29.95 &     16.72 &    13.47 &  10.45 \\
       FISTA large $\mu$ &     1.04 &     0.97 &      1.12 &      1.14 &       --- &      --- &    --- \\
    \bottomrule
    \end{tabular}
    \label{tab:algos-convergence_mri}
\end{table}

% FISTA with fixed $\mu$
\figref{fig:algos-convergence_mri} and \tabref{tab:algos-convergence_mri} shows that FISTA with fixed $\mu$ is either too slow (Chen's $\mu$) or, as expected, does not reach the desired precision (as with the large $\mu$).
However, CONESTA compete with FISTA with large $\mu$ and EGM during the first iterations. This demonstrates two points: CONESTA dynamically picked an efficient (large enough) smoothing parameter and that the gap stopping criterion, used in the nested FISTA loop, allows us to stop before reaching a plateau, and thus to quickly change to a smaller $\mu$ (illustrated with dots in \figref{fig:algos-convergence_mri} where the continuation steps occurred).

% Inexact slow down over FISTA loop
\figref{fig:algos-convergence_mri}, bottom panel, shows the fast convergence of Inexact FISTA as a function of the number of iterations.
However, the top panel of \figref{fig:algos-convergence_mri} shows that it is always considerably slower, in terms of the execution time, compared to the EGM or CONESTA.
\tabref{tab:algos-convergence_mri} reveals that Inexact FISTA is 4.38 times slower than CONESTA to reach an error of $10^{-3}$ and this difference in speed increases with higher precisions.
This demonstrates the hypothesis we stated in the introduction, that Inexact FISTA becomes slower after many iterations due to the necessity to decrease the precision faster than $1/k^4$ ($k$ being the number of FISTA iterations), in the approximation.
%We experimentally observed the following phenomenon: during the first 100 FISTA iterations, only a few ($<10$) iterations of the nested loop was necessary, but then this number increased progressively and exceeded 100 after a few thousand FISTA iterations.

As a conclusion, \figref{fig:algos-convergence_mri} and \tabref{tab:algos-convergence_mri} illustrate that on high-dimensional MRI data sets, CONESTA outperformed all other algorithms for precisions higher than $\varepsilon \leq 10^{-2}$.

\begin{figure}[!h]
     \centering
     \includegraphics[width=0.5\textwidth]{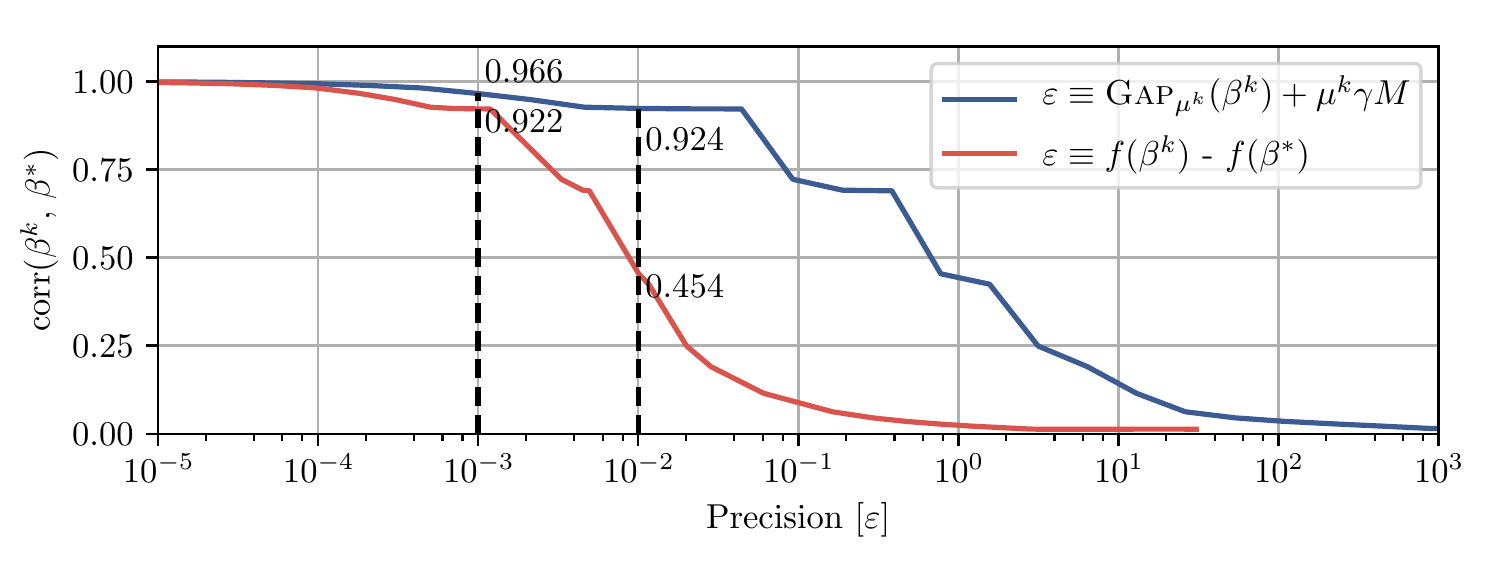}\\
     \includegraphics[width=0.5\textwidth]{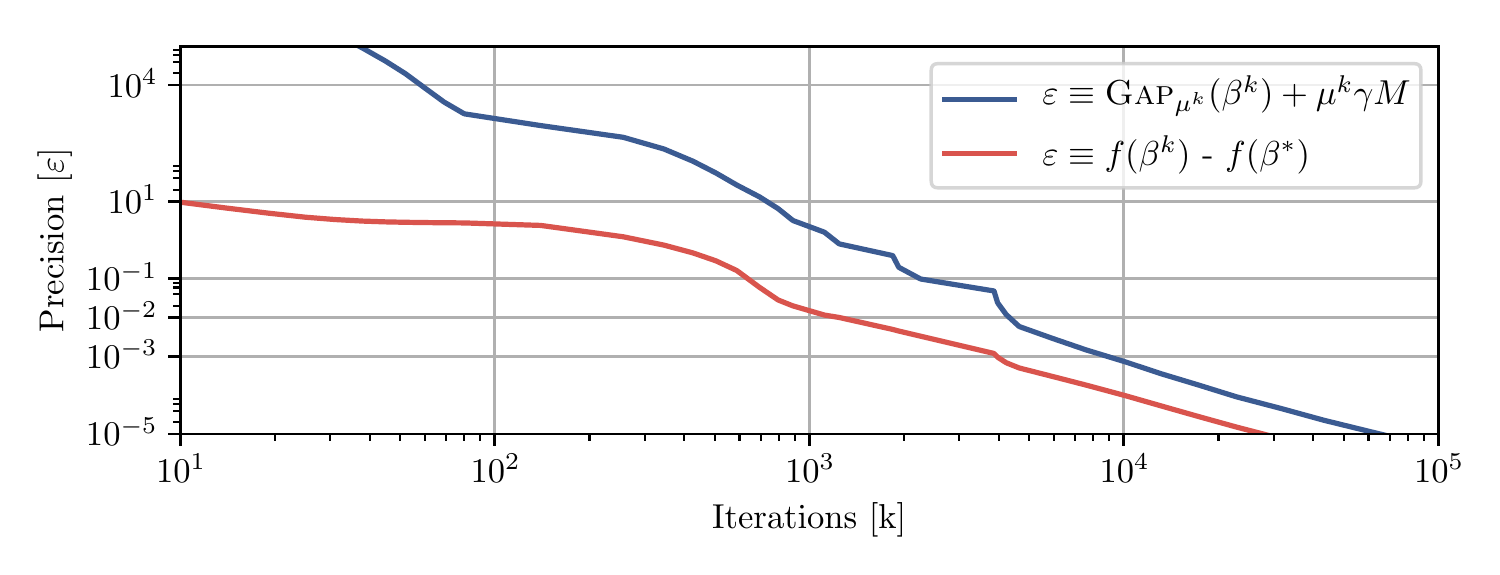}
     \caption{\FIRST{Top panel: Correlation between the coefficient maps $\bbeta^k$ and the true solution $\bbeta^*$ as a function of the true precision (red line) and precision estimated with the duality gap.
              The true solution has been estimated by running $10^6$ iterations of CONESTA and Inexact FISTA.
              %The dashed line indicates that indicates that a precision of $10^{-3}$ is required to obtain a coefficients maps that is very similar to the true solution.
	      Bottom panel: True precision (red line) and precision estimated with the duality gap (blue line) as a function of the number of iteration.
	      $10^4$ iterations are sufficient to reach and outperform the required precision of $10^{-3}$.}}
\label{fig:precision_gap}
\end{figure}

\FIRST{
\subsubsection{Required precision and its gap estimate}\label{sec:required-precision}

The figure \figref{fig:precision_gap}, top panel, indicates that for precisions higher than $10^{-3}$ (using both the duality gap estimation of the precision and the true precision), the similarity of coefficient maps to the true solution is reaching a plateau.
An early stopping at $\varepsilon=10^{-2}$ would provide a different solution than the expected one using either measures of precision: $\text{corr}(\bbeta^k, \bbeta^*)=0.92$ with the gap estimate and $\text{corr}(\bbeta^k, \bbeta^*)=0.45$ with the true precision.
Moreover the figure demonstrates the relevance of the duality gap as a stopping criterion: stopping the convergence at $10^{-3}$, using the duality gap, provides a map with a $0.97$ correlation with the true solution.
The bottom panel shows that less than $10^4$ iterations are sufficient to reach the target precision of $10^{-3}$, which is less than 30 minutes of computation.
It also shows that, for useful precisions ($\varepsilon \leq 10^{-1}$), the duality gap is an accurate upper-bound of the true error.
}

%%%%%%%%%%%%%%%%%%%%%%%%%%%%%%%%%%%%%%%%%%%%%%%%%%%%%%%%%%%%%%%%%%%%%%%%%%%%%%%%
\section{Conclusion}
%%%%%%%%%%%%%%%%%%%%%%%%%%%%%%%%%%%%%%%%%%%%%%%%%%%%%%%%%%%%%%%%%%%%%%%%%%%%%%%%

We investigated in this paper an optimization problem where a linear regression model was combined with several convex penalties, including a structured penalty under very general assumptions on the structure and on the smooth loss function.

We proposed an algorithm, CONESTA, that aims to overcome a set of obstacles from current state-of-the-art algorithms, notably non-separability and non-smoothness of penalties.
Our results show that the proposed algorithm possesses both desirable theoretical convergence guarantees and practical scalability properties under various settings involving complex structured constraints and high dimensionality.
The robustness and scalability of CONESTA is a key feature to processing high-dimensional whole-brain neuroimaging data.
Specifically, the main benefits of CONESTA compared to state-of-the-art methods can be summarized in the following points:

\begin{enumerate}
    \item CONESTA has a convergence rate of $\bigO{1/\varepsilon}$, an improvement over FISTA with fixed smoothing whose rate is $\bigO{1/\varepsilon} + \bigO{1/{\sqrt{\varepsilon}}}$~\cite{teb2, Chen2012}.

    \item CONESTA outperformed (in terms of execution time and precision of the solution) several state-of-the-art optimization algorithms on both simulated and neuroimaging data.

    \item CONESTA is a robust solver that resolves practical problems affecting other solvers in very high dimensions encountered in neuroimaging. For instance:
    (i) The EGM does not allow true sparsity, nor complex loss functions.
    (ii) The Inexact FISTA converges faster (in terms of the number of iterations) compared to CONESTA. However, as observed on the high-dimensional MRI dataset, after hundreds of iterations, solving the subproblem using an inner FISTA loop makes it much slower (\eg, \FIRST{it took} 4 times \FIRST{longer} to reach $\varepsilon<10^{-3}$) compared to CONESTA.
    (iii) ADMM is often not robust to the exact choice of the parameters, and does not scale well to arbitrary complex penalties or \eg 3D images. Conversely, CONESTA requires only a global precision and, thanks to the duality gap, it will dynamically adapt the decreasing smoothing sequence.

    \item The algorithm that we propose is able to include any combination of terms from a variety of smooth and non-smooth penalties with smooth losses.
    For example, in \cite{de_pierrefeu_structured_2017} we exploited the versatility of CONESTA as a building block to solve a Principal Components Analysis problem with both TV and elastic-net penalties. 
    CONESTA does not require the proximal operators or any other auxiliary minimization problem, contrary to many other similar state-of-the-art algorithms. For instance, any convex penalty in the form of a sum of $\ell_p$-norms of linear operators applied on all or on a subset of all variables could be used with this algorithm together with any smooth loss function for which we know the gradient.

    \item As a result of the proposed duality gap, CONESTA has a rigorous stopping criterion that allows a desired precision to be reached with theoretical certainty.
\end{enumerate}

Finally, CONESTA has been designed to work well with warm restarts. Indeed, the whole smoothing sequence is dynamically adapted to the distance to the global minimum.
This is a crucial feature for a solver that handles high-dimensional neuroimaging problems, where the user will explore a grid of possible penalty parameters.
Since a first solution obtained with a set of parameter is similar to that sought with a slight modification of a parameter, the first solution will be an efficient starting point to quickly find the solution to the second problem.

%Finally, we note again that CONESTA easily could be adapted to many other penalties. For example, to use the overlapping group lasso (GL) penalty in our framework, we just have to design a particular linear operator, $\A_{GL}$, and use as the linear operator $\A$ in \algref{algo:conesta}.

% if have a single appendix:
%\appendix[Proof of the Zonklar Equations]
% or
%\appendix  % for no appendix heading
% do not use \section anymore after \appendix, only \section*
% is possibly needed
%\comment{mention that we used pylearn-parsimony for the optimizations, and give reference/link}

\appendix[ParsimonY: structured and sparse machine learning in Python]\label{sec:parsimony}

The algorithms investigated here are provided in the ParsimonY \FIRST{Python } library \FIRST{found at} \url{https://github.com/neurospin/pylearn-parsimony}.
\FIRST{ParsimonY is a Python library that intends to be compliant with the scikit-learn machine learning library, and provides efficient solvers for a very general class of optimization problems including many group-wise penalties (allowing overlapping groups) such as Group Lasso and TV.}
Listing \ref{list:parsimony} shows \FIRST{an implementation of the solution to a problem with} elastic net and TV (default solver is CONESTA) assuming a data set \texttt{X} of size $199\times 286\,217$ where a row contains the age, gender, education and the $286\,214$ GM voxels of each of the $199$ patients.
Here we left the first three demographic columns un-penalized (\texttt{penalty\_start=3}), (defaults is 0, all columns are penalized).
The neuroimaging dataset can be found at \url{ftp://ftp.cea.fr/pub/unati/brainomics/papers/ols_nestv}.

%\usepackage{listings}
%\lstset{basicstyle=\ttfamily}
\lstset{%language=python,
    basicstyle=\footnotesize\ttfamily,   
    keywordstyle=\bfseries,
    morecomment=[l]{\#},
    %commentstyle=\footnotesize\ttfamily,   
    showstringspaces=false,
    morekeywords={fit, predict, beta, LinearRegressionL1L2TV, linear_operator_from_mask}
}

%{\footnotesize
\begin{lstlisting}[caption=Elastic net and TV with the ParsimonY library, label=list:parsimony]
import parsimony.functions.nesterov.tv as tv
from parsimony.estimators import LinearRegressionL1L2TV

mask_ima = nibabel.load("mask.nii")
Atv = tv.linear_operator_from_mask(mask_ima.get_data())
mod = LinearRegressionL1L2TV(l1=0.33, l2=0.33, ltv=0.33,
    A=Atv, penalty_start=3)
mod.fit(X, y)  # Fit the model
mod.predict(X)  # Predict scores
\end{lstlisting}
%}
\iffalse

# Save weights map into nifti image
weight_arr = np.zeros(mask_ima.get_data().shape)
weight_arr[mask_ima.get_data() != 0] = \
  mod.beta.ravel()[penalty_start:]
weight_nii = nibabel.Nifti1Image(weight_arr,
  affine=mask_ima.get_affine())
weight_nii.to_filename("weights.nii")

% use section* for acknowledgement
\section*{Acknowledgment}

The authors would like to thank...
\fi

%%%%%%%%%%%%%%%%%%%%%%%%%%%%%%%%%%%%%%%%%%%%%%%%%%%%%%%%%%%%%%%%%%%%%%%%%%%%%%%%
%%                               REFERENCES                                   %%
%%%%%%%%%%%%%%%%%%%%%%%%%%%%%%%%%%%%%%%%%%%%%%%%%%%%%%%%%%%%%%%%%%%%%%%%%%%%%%%%

\bibliographystyle{plain}
\bibliography{references}
\end{document}